\documentclass{article} 
\usepackage{iclr2027_conference,times}

\usepackage{amsmath,amsfonts,bm}

\def\eqref#1{equation~\ref{#1}}

\def\1{\bm{1}}

\DeclareMathAlphabet{\mathsfit}{\encodingdefault}{\sfdefault}{m}{sl}
\SetMathAlphabet{\mathsfit}{bold}{\encodingdefault}{\sfdefault}{bx}{n}

\usepackage{hyperref}
\usepackage{url}
\usepackage{graphicx}

\usepackage{bm}
\renewcommand{\vec}[1]{\bm{#1}}

\title{A theory of noisy linear recurrent \\neural network learning dynamics}

\author{%
  John J. Vastola, Kanaka Rajan \\
Department of Neurobiology, Harvard Medical School \\
Kempner Institute for the Study of Natural and Artificial Intelligence, Harvard University \\
  \texttt{\{john\_vastola, kanaka\_rajan\}@hms.harvard.edu} \\
}

\iclrfinalcopy 
\begin{document}

\maketitle
\pagestyle{plain}

\begin{abstract}
Biological neural networks are noisy: their dynamics involve difficult-to-predict fluctuations which are likely due not just to chaos, but also to a variety of low-level biophysical processes. Despite the omnipresence of intrinsic noise, its theoretical consequences are incompletely understood, especially in the context of learning. For example, does noise meaningfully shape representation learning, or is it better viewed as a nuisance factor whose presence or absence changes little about what solution is found and when? Here, we argue that intrinsic noise can significantly modulate representation learning in recurrent neural networks (RNNs), since it can strongly break solution-space degeneracies and in some cases control whether a non-degenerate solution exists at all. We show that this is true for an analytically tractable model of noisy RNN learning with a linear activation function and white noise inputs. We find that, in the so-called rich regime (where weights are initialized with small values), noise strongly modulates the directions in which the read-in, readout, and recurrent weights align. Moreover, we find that noise produces a kind of phase transition: for sufficiently high noise, a non-degenerate solution is not even optimal if parameter scales are limited (e.g., via regularization). These results extend recent analytic work on RNN representation learning and show that noise can act as a fundamental determinant of representations.
\end{abstract}

\section{Introduction}

Recurrent neural networks (RNNs) are extraordinarily successful models of the brain \citep{vogels_neural_2005,mante_context-dependent_2013,rajan_recurrent_2016,vyas_computation_2020}, and the past two decades of theory and experiment have produced a working understanding of what they learn and how. Their loss landscapes are generally more complex than those of feedforward networks, in part because recurrence can produce exploding and vanishing gradients \citep{bengio_explode_1994,pascanu_exploding_2013}; their solution space is generally highly degenerate \citep{huang_measuring_2025}; and they can be biased towards learning low-rank solutions \citep{schuessler_interplay_2020}. But a major feature of RNNs that remains poorly understood is the significance of \textit{intrinsic noise}. 

Noise is omnipresent in biological neural networks: spike trains and firing rate traces throughout the nervous system involve difficult-to-predict fluctuations \citep{faisal_noise_2008}, which may be a manifestation of chaotic behavior \citep{sompolinsky_chaos_1988,crisanti_chaos_2018} or low-level stochasticity associated with chemical reactions. Although from one perspective noise is a nuisance, from another it provides pressure for networks to find more robust solutions. Noise is routinely used in RNN training \citep{yang_task_2019,driscoll_flexible_2024}, in part to improve generalization, and there is evidence that noise can drive networks into flatter loss landscape minima \citep{lim_noisy_2021}.

From a representation-learning perspective, it may be helpful to view noise as having two potential consequences. First, noise can provide \textit{geometric constraints} on solutions: weights must be oriented in certain directions (i.e., the less noisy directions) and not others, and their overall scales must be chosen to minimize noise. Second, the goal of minimizing noise may compete with that of achieving good task performance; if there is too much noise, one expects that the system may be driven to a trivial solution at which task performance is poor, but there is little or no noise.

Although RNN learning dynamics are difficult to study, an interesting toy model of them has emerged in recent work: \textit{linear} RNNs, especially linear RNNs trained to perform leaky integration of a white noise stimulus \citep{proca_rnn_2025,bordelon_rnn_2025}. As \cite{bordelon_rnn_2025} show, when network parameters are initially small, learning has two phases: an \textit{alignment phase}, in which parameter movement concentrates along a specific (initialization-dependent) direction of state space, followed by a \textit{dynamics phase}, in which an eigenvalue `escapes' from the bulk in order to solve the task.

In this work, we consider an extremely similar theoretical setup, whose primary difference is the presence of (generally anisotropic) intrinsic noise. We show that their theory must be substantially modified even if there is only a small amount of noise, since noise affects both the alignment and dynamics phases, and even whether a nontrivial solution exists at all. Throughout, we focus on characterizing learning in the \textit{rich regime}, as opposed to the \textit{lazy regime}, in part because the rich regime is thought to be more representative of learning in performant RNNs.

Our main contributions are as follows.
\begin{itemize}
\item We develop an analytically tractable theory of representation learning in linear RNNs with intrinsic noise. We show that \textbf{noise breaks degeneracies of the original objective}, which yields preferred state space directions and read-in/readout scales.
\item We characterize how intrinsic noise modifies rich-regime learning dynamics. During the alignment phase, anisotropic noise changes the growth rates of different state-space modes, which \textbf{biases learning toward less noisy directions}. Noise also breaks the symmetry between the read-in and readout weights, and in the absence of regularization, drives the input weight scale to infinity and the output weight scale to zero.
\item We show that \textbf{sufficiently strong noise can produce phase transitions beyond which only a trivial solution is learned}. This is true both when all parameters are trainable and when some parameters are frozen.
\end{itemize}

\section{Mathematical formulation}
\label{sec:math}

\begin{figure}[ht!]
\begin{center}
\includegraphics[width=\textwidth]{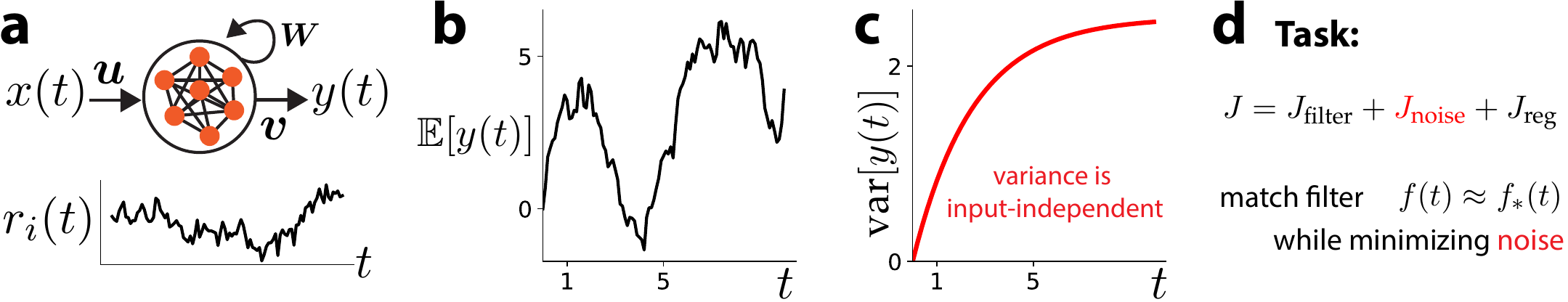}
\end{center}
\caption{\textbf{Theoretical setting.} (\textbf{a}) We study a noisy linear RNN which receives a scalar white noise signal $x(t)$ as input and produces an output $y(t)$. The dynamics of individual $r_i$ are noisy (bottom). (\textbf{b}) On average, the network filters its input. (\textbf{c}) Output variance is input-independent and eventually saturates (assuming the network is stable). (\textbf{d}) We assume the network is trained to optimize an objective with three terms: a filter-matching term that incentivizes matching its average behavior to that of a target filter $f_*(t)$,  a noise reduction term, and a regularization term.}
\label{fig:setup}
\end{figure}

Our setup largely mirrors that of \cite{bordelon_rnn_2025}. Throughout this paper, we study a linear RNN with $N \geq 1$ neurons whose instantaneous state is characterized by a vector $\vec{r} := (r_1, ..., r_N)^T \in \mathbb{R}^N$. We assume dynamics
\begin{equation} \label{eq:dyn}
\dot{\vec{r}} = \vec{A} \vec{r} + x(t) \vec{u} + \vec{G} \ \vec{\eta}(t) = - \gamma \vec{r} + \vec{W} \vec{r} + x(t) \vec{u} + \vec{G} \ \vec{\eta}(t)
\end{equation}
where $x(t)$ is a time-dependent input signal, $\vec{u} \in \mathbb{R}^N$ are the \textit{read-in weights}, $\gamma > 0$ is the coefficient of the \textit{leak term}, and $\vec{W} \in \mathbb{R}^{N \times N}$ are the \textit{recurrent weights} (Fig. \ref{fig:setup}a). To ease notation, we sometimes work with the combination $\vec{A} := - \gamma \vec{I} + \vec{W}$. The remaining term instantiates \textit{intrinsic noise}: $\vec{G} \in \mathbb{R}^{N \times N}$ controls its amplitude, and $\vec{\eta}(t) \in \mathbb{R}^N$ is a vector of Gaussian white noise processes. We are most interested in understanding how the structure of $\vec{G}$ shapes learning.

Because Eq. \ref{eq:dyn} is linear in $\vec{r}$ and involves only state-independent additive noise, we can immediately write down the distribution of $\vec{r}(t)$ conditional on any initial condition, although we will assume $\vec{r}(0) = \vec{0}$ for simplicity. We find that $\vec{r}(t)$ is normally distributed with
\begin{equation}
p(\vec{r} , t | \vec{r}_0 = \vec{0}, 0) = \mathcal{N}\left(  \int_0^t e^{\vec{A} t'} \vec{u} \ x(t-t') \ dt'    ,   \int_0^t  e^{\vec{A} t'} \vec{S} e^{\vec{A}^T t'} \ dt'  \right) 
\end{equation}
where $\vec{S} := \vec{G} \vec{G}^T$ is, up to a factor of $2$, the \textit{diffusion tensor} associated with $\vec{G}$ (which we take to be positive definite). On average, the network linearly filters its input (Fig. \ref{fig:setup}b). But because this integration is noisy, $\vec{S}$-dependent noise accumulates until (provided the eigenvalues of $\vec{A}$ all have negative real parts) it saturates (Fig. \ref{fig:setup}c).

\subsection{Objective function}

Following \cite{bordelon_rnn_2025}, assume a linear readout determined by \textit{readout weights} $\vec{v} \in \mathbb{R}^N$:
\begin{equation}
y(t) := \vec{v}^T \vec{r}(t) \ .
\end{equation}
This readout defines a filter $f(t) := e^{- \gamma t} \vec{v}^T e^{\vec{W} t} \vec{u}$ since, on average,
\begin{equation}
\mathbb{E}_{\vec{r}}[ y(t) ] = \int_0^t \vec{v}^T e^{\vec{A} t'} \vec{u} \ x(t-t') \ dt' = \int_0^t f(t') \ x(t-t') \ dt' \ .
\end{equation}
Assume we train some combination of $\vec{u}$, $\vec{v}$, and $\vec{W}$ so that $y(t)$ matches a \textit{target output}
\begin{equation}
y_*(t) := \int_0^t e^{- c_* t'} x(t - t') \ dt'  = \int_0^t f_*(t') x(t - t') \ dt'
\end{equation}
which linearly filters its input $x(t)$ using the \textit{target filter} $f_*(t) := e^{- c_* t}$. Here, $c_* > 0$ controls the time scale on which input is leaky-integrated. Assume we do this using the supervised objective
\begin{equation}
J := \lim_{T \to \infty} \frac{1}{T} \int_0^T \frac{1}{2} \mathbb{E}_{x, \vec{r}}\left\{ \ [ y(t) - y_*(t) ]^2 \ \right\} \ dt + \frac{\kappa}{2} \Vert \vec{u} \Vert^2  + \frac{\kappa}{2} \Vert \vec{v} \Vert^2  \ ,
\end{equation}
which demands that the network output $y(t)$ agrees with $y_*(t)$ on average. Note that we have also included a regularization term whose strength is determined by $\kappa \geq 0$. This objective can be simplified if the input has \textit{white noise} statistics \citep{proca_rnn_2025,bordelon_rnn_2025}, i.e., if $\mathbb{E}_x[ x(t) ] = 0$ and $\mathbb{E}_x[ x(t) x(t - t') ] = \delta(t - t')$. We find (see Appendix \ref{app:obj})
\begin{equation}
\boxed{ \ 
J(\vec{u}, \vec{v}, \vec{W}) =  \int_0^{\infty}    \frac{1}{2} \left[ f(t) - f_*(t) \right]^2 +  \frac{e^{- 2 \gamma t}}{2}  \vec{v}^T e^{\vec{W} t} \vec{S} e^{\vec{W}^T t} \vec{v} \   dt   + \frac{\kappa}{2} \Vert \vec{u} \Vert^2  + \frac{\kappa}{2} \Vert \vec{v} \Vert^2 \ } \ .
\end{equation}
Note that $J$ has three terms: (i) a \textit{filter-matching term} $J_{\text{filter}}$, (ii) a \textit{noise term} $J_{\text{noise}}$, and (iii) a \textit{regularization term} $J_{\text{reg}}$ (Fig. \ref{fig:setup}d). \cite{bordelon_rnn_2025} only considered the filter-matching term; we will see that the other two terms strongly modulate learning dynamics.

\subsection{Learning dynamics}

RNNs are normally trained via backpropagation through time \citep{werbos_bptt_1990}, FORCE \citep{sussillo_generating_2009}, or real-time recurrent learning \citep{williams_rtrl_1989,irie_exploring_2024}. One virtue of linear RNNs, at least for theory, is that gradients of $J$ with respect to $\vec{u}$, $\vec{v}$, and $\vec{W}$ can be computed exactly, so we can ignore the extra structure associated with those learning rules. Assuming parameters are updated via gradient descent, in the gradient-flow limit in which the learning rate is taken infinitesimally small, we recover ordinary differential equations (ODEs) (see Appendix \ref{app:obj})
\begin{equation} \label{eq:dyn_learn}
\begin{split}
\frac{d}{d\tau} \vec{u} &=  \int_0^{\infty} [ f_*(t) - f(t) ] e^{- \gamma t} e^{\vec{W}^T t} \vec{v}  \ dt - \kappa \vec{u}  \\
\frac{d}{d\tau} \vec{v} &= \int_0^{\infty} [ f_*(t) - f(t) ] e^{- \gamma t} e^{\vec{W} t} \vec{u}  -  e^{- 2 \gamma t} e^{ \vec{W} t} \vec{S} e^{ \vec{W}^T t} \vec{v}  \  dt - \kappa \vec{v} \\
\frac{d}{d\tau} \vec{W} &= \int_0^{\infty} [ f_*(t) - f(t) ] e^{- \gamma t} \int_0^t  e^{\vec{W}^T t'} \vec{v} \vec{u}^T e^{\vec{W}^T (t - t')} \ dt' dt\\
&- \int_0^{\infty} e^{-2\gamma t} \int_0^t \  e^{\vec{W}^T t'} \vec{v} \vec{v}^T e^{\vec{W} t} \vec{S} e^{\vec{W}^T (t - t')} \ dt' dt 
\end{split}
\end{equation}
parameterized by a real-valued `learning time' $\tau \in [0, \infty)$. These dynamics are generically highly nonlinear and high-dimensional, although we will see shortly that they simplify substantially if the initial weights are assumed to be small.


\section{Noise produces representational pressure}

\begin{figure}[ht!]
\begin{center}
\includegraphics[width=\textwidth]{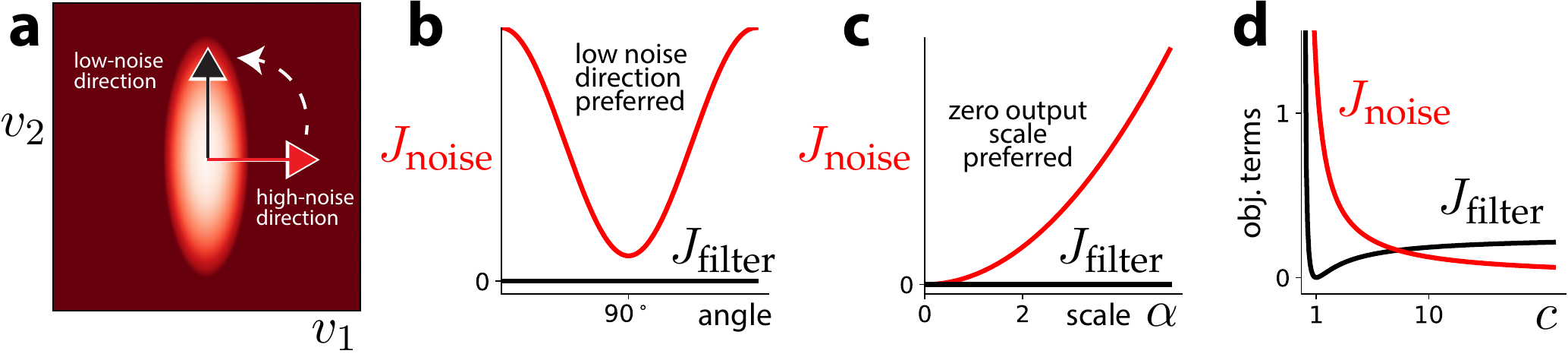}
\end{center}
\caption{\textbf{Noise produces representational pressure.} (\textbf{a}) Two representations with the same filter but different readout vectors differ in noise level if $\vec{S}$ is anisotropic. Pictured is a schematic 2D example with a high-noise direction (horizontal) and a low-noise direction (vertical). One can reduce noise by rotating $\vec{v}$ (white arrow). (\textbf{b}) The noise term $J_{\text{noise}}$ in the schematic example as a function of $\vec{v}$ orientation. $J_{\text{filter}}$ is unaffected as long as the filter remains the same. (\textbf{c}) $J_{\text{noise}}$ and $J_{\text{filter}}$ as a function of $\alpha$ (see main text); a zero readout scale is preferred given noise. (\textbf{d}) $J_{\text{filter}}$ and $J_{\text{noise}}$ vs $c$ (see main text) generally have different optima, and their competition can change where the minimum of $J$ is, or whether it has a finite minimum at all.}
\label{fig:pressure}
\end{figure}

One elemental observation is that noise produces \textit{representational pressure}: whereas the noiseless system has many representations compatible with good performance, the noisy system must minimize the corrupting influence of noise, which can greatly reduce the number of good solutions.

\paragraph{The deterministic solution is highly degenerate.}

In order for $f(t) = e^{- \gamma t} \vec{v}^T e^{\vec{W} t} \vec{u}$ to match $f_*(t) = e^{- c_* t}$, assuming $\vec{W}$ is diagonalizable, it must be true that (i) $\gamma - c_*$ is an eigenvalue of $\vec{W}$; (ii) $\vec{u}$ has a component along one of the corresponding \textit{right} eigenvectors; (iii) $\vec{v}$ has a component along one of the corresponding \textit{left} eigenvectors; and (iv) $\vec{u}^T \vec{v} = 1$. But this yields a solution space with huge parameter degeneracies, including invariance to arbitrary invertible linear transformations. In particular, if $\vec{T} \in \mathbb{R}^{N \times N}$ is an invertible linear transformation, then $f(t)$ remains invariant to the mapping with $\vec{v} \to \vec{T}^T \vec{v}$, $\vec{u} \to \vec{T}^{-1} \vec{u}$, and $\vec{W} \to \vec{T}^{-1} \vec{W} \vec{T}$, since
\begin{equation}
\vec{v}^T e^{\vec{W} t} \vec{u} \xrightarrow{\vec{T}} \vec{v}^T \vec{T} \vec{T}^{-1} e^{\vec{W} t} \vec{T} \vec{T}^{-1} \vec{u} = \vec{v}^T e^{\vec{W} t} \vec{u} \ .
\end{equation}
Interesting special cases include when $\vec{T}$ is a rotation or a change in read-in/readout scale. Meanwhile, the noise term in our objective does \textit{not} generally respect those degeneracies:
\begin{equation}
\int_0^{\infty} e^{- 2 \gamma t} \vec{v}^T e^{\vec{W} t} \vec{S} e^{\vec{W}^T t} \vec{v} \ dt \xrightarrow{\vec{T}}   \int_0^{\infty} e^{- 2 \gamma t} \vec{v}^T e^{\vec{W} t} (\vec{T} \vec{S}  \vec{T}^{T}) e^{\vec{W}^T t}  \vec{v} \ dt   \ .
\end{equation}
We discuss various interesting consequences of this simple observation below.

\paragraph{Noise picks out preferred state-space directions.}

If multiple state-space directions could solve the task (i.e., if $\vec{W}$ has multiple correct eigenvalues), the presence of noise causes \textit{less noisy} directions to be preferred (Fig. \ref{fig:pressure}a, b). For example, if $\vec{W} = (\gamma - c_*) \vec{I}$ (i.e., all eigenvalues are correct) and $\vec{v}^T \vec{u} = 1$ (so that $f(t) = f_*(t)$), then the noise term is minimized when
\begin{equation}
\vec{v}_* = \frac{\vec{S}^{-1} \vec{u}}{\vec{u}^T \vec{S}^{-1} \vec{u}} \ ,
\end{equation}
i.e., when state-space directions are read out according to how noisy they are, with $\vec{v}_*$ aligning more with less noisy directions. Essentially this solution is found when only the readout $\vec{v}$ is trained, i.e., when a `reservoir computing' approach \citep{LUKOSEVICIUS_reservoir_2009} is used (see Appendix \ref{app:reservoir}).

\paragraph{Noise produces preferred read-in and readout scales.}

A trivial way to reduce the noise term is to take the scale of the readout weights to zero. This is consistent with the above general observation; if $\vec{T} = \alpha \vec{I}$, the filter-matching term remains unchanged but the noise term picks up a factor of $\alpha^2$, and the overall term is minimized when we take $\alpha \to 0$ (Fig. \ref{fig:pressure}c). This implies, given no regularization ($\kappa = 0$), that it is optimal to take the readout scale to zero and the read-in scale to infinity.

\paragraph{Noise minimization can compete with filter-matching accuracy.} The filter-matching term $J_{\text{filter}}$ and the noise term $J_{\text{noise}}$ generally do not share optima, so their relative size can determine where the fixed points of learning dynamics are, or whether there is any finite fixed point at all. Fig. \ref{fig:pressure}d depicts an example we will consider later in Sec. \ref{sec:ps2}: the balance of the two terms determines whether the recurrent weight scale $c$ converges to a finite value near $c_*$ or diverges to infinity.

\section{Noise determines weight alignment in rich regime}
\label{sec:align}

\begin{figure}[ht!]
\begin{center}
\includegraphics[width=\textwidth]{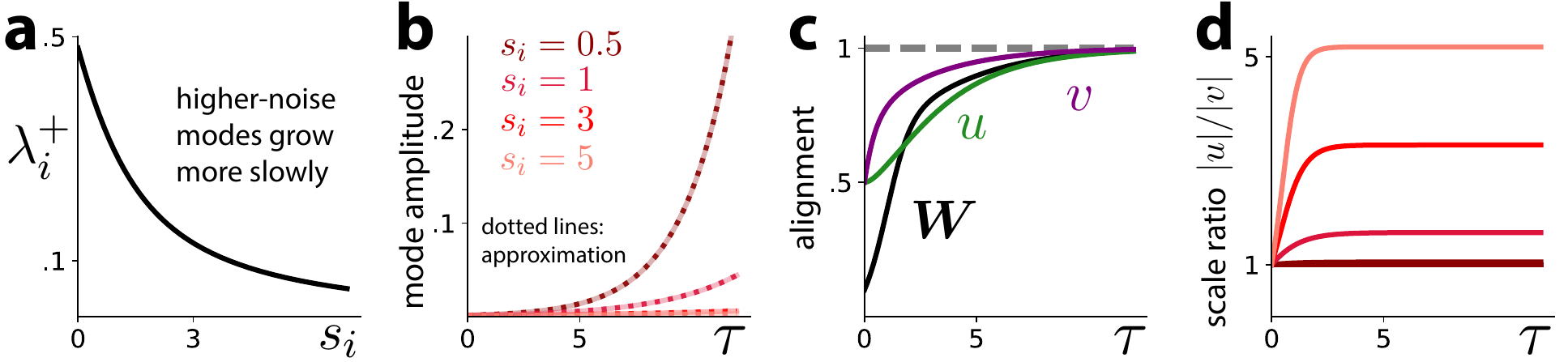}
\end{center}
\caption{\textbf{Noise determines alignment.} (\textbf{a}) The dominant contribution to $u_i(\tau)$ and $v_i(\tau)$ grows exponentially quickly at a rate $\lambda^+_i$ that decreases as the corresponding $\vec{S}$ eigenvalue (i.e., noise level) $s_i$ increases, so higher-noise modes grow more slowly. (\textbf{b}) Although dynamics involves both growing and decaying contributions, after enough learning time has elapsed, mode amplitudes (solid lines) are well-approximated by only their growing contribution (dashed lines). (\textbf{c}) The dot product between $\vec{u}$ and $\vec{v}$, and the least noisy eigenvector of $\vec{S}$, increases during this phase. $\vec{W}$ aligns in the same direction and becomes approximately rank-one. (\textbf{d}) The discrepancy between the scales of $\vec{u}$ and $\vec{v}$ grows with learning, and is higher in directions with larger noise levels. }
\label{fig:align}
\end{figure}

In this section, we begin our analysis of this model's learning dynamics (Eq. \ref{eq:dyn_learn}). We focus on the so-called \textit{rich regime}, in which weights are initialized to small values and low-rank structure emerges. Our analysis here follows the spirit of \cite{bordelon_rnn_2025} but differs in its details.

\subsection{Dynamics given small initial weights}

Assume $\vec{u}$, $\vec{v}$, and $\vec{W}$ are initialized with small values (say, of order $\epsilon > 0$). To leading order in $\epsilon$, gradient descent learning dynamics become (see Appendix \ref{app:obj})
\begin{equation} \label{eq:align}
\frac{d}{d\tau} \vec{u} \approx  \frac{\vec{v} }{c_* + \gamma} - \kappa \vec{u} \hspace{0.28in} \frac{d}{d\tau} \vec{v} \approx \frac{\vec{u}}{c_* + \gamma}    -  \frac{\vec{S} \vec{v}}{2 \gamma} - \kappa \vec{v} \hspace{0.28in}  \frac{d}{d\tau} \vec{W} \approx \frac{\vec{v} \vec{u}^T }{(c_* + \gamma)^2}  - \frac{ \vec{v} \vec{v}^T \vec{S} }{4 \gamma^2}  \ .
\end{equation}
In other words, $\vec{u}$ and $\vec{v}$ tend to evolve according to a certain linear system, while $\vec{W}$ depends only on input from $\vec{u}$ and $\vec{v}$. This is a major improvement, since Eq. \ref{eq:dyn_learn} is in general a highly nonlinear system of ODEs. Note that the presence of $\vec{S}$ yields dynamics different from those studied by \cite{bordelon_rnn_2025}; $\vec{S}$ breaks the symmetry between $\vec{u}$ and $\vec{v}$ dynamics, since components of $\vec{v}$ along high-noise eigenvectors of $\vec{S}$ are penalized while the analogous components of $\vec{u}$ are not.

\subsection{Weights rotate away from highly noisy state space directions}

The alignment-phase ODEs (Eq. \ref{eq:align}) simplify if we use the eigenvectors of $\vec{S}$ as a basis. Letting $u_i$ and $v_i$ denote the components of $\vec{u}$ and $\vec{v}$ along an eigenvector whose eigenvalue is $s_i$, we have
\begin{equation}
\frac{d}{d\tau} u_i = - \kappa u_i + \frac{v_i}{c_* + \gamma}  \hspace{1in} \frac{d}{d\tau} v_i = \frac{u_i}{c_* + \gamma}    - \left( \kappa + \frac{s_i}{2 \gamma} \right) v_i \ ,
\end{equation}
i.e., components along different eigenvectors evolve independently. The $2 \times 2$ matrix that governs these dynamics has two eigenvalues (see Appendix \ref{app:align}):
\begin{equation}
\lambda^{\pm}_i := - \left(  \kappa + \frac{s_i}{4 \gamma}    \right) \pm \sqrt{ \frac{s_i^2}{16 \gamma^2}    + \frac{1}{(c_* + \gamma)^2} } \ .
\end{equation}
If the regularization strength $\kappa$ is too large, both eigenvalues are negative, and $u_i$ and $v_i$ collapse to a trivial zero solution. Otherwise, $\lambda^+_i$ is positive and $\lambda_i^-$ is negative, which means that the $\lambda^+_i$ component of the solution tends to dominate after an initial transient. More quantitatively,
\begin{equation}
\begin{split}
u_i(\tau) &\approx  \frac{1}{\lambda^+_i - \lambda^-_i} \left[ \frac{v_i(0)}{c_* + \gamma} - u_i(0) (\kappa + \lambda^-_i) \right] e^{\lambda^+_i \tau} \\
v_i(\tau) &\approx  \frac{(c_* + \gamma) (\kappa + \lambda^+) }{\lambda^+_i - \lambda^-_i} \left[ \frac{v_i(0)}{c_* + \gamma} - u_i(0) (\kappa + \lambda^-_i) \right] e^{\lambda^+_i \tau} \ .
\end{split}
\end{equation}
One interesting feature of this result is that the growth rate $\lambda^+_i$ is a \textit{decreasing} function of $s_i$ (Fig. \ref{fig:align}a), which implies that components of $\vec{u}$ and $\vec{v}$ in noisier directions grow more slowly (Fig. \ref{fig:align}b). Because small differences are exponentially amplified, to good approximation $\vec{u}$ and $\vec{v}$ align along the \textit{least noisy} eigenvectors of $\vec{S}$ (Fig. \ref{fig:align}c). In particular,
\begin{equation} \label{eq:align_result}
\begin{split} 
\vec{u}(\tau) &\approx u(\tau) \vec{P}  \left[  \frac{[ \vec{u}(0) + \vec{v}(0) ]}{2} +  \frac{(c_* + \gamma)}{2 } \left(\frac{s_{\text{min}}}{4 \gamma}  + \Delta \right) \vec{u}(0)    \right]  \\
\vec{v}(\tau) &\approx v(\tau) \vec{P}   \left[  \frac{[ \vec{u}(0) + \vec{v}(0) ]}{2} + \frac{(c_* + \gamma)}{2 } \left(\frac{s_{\text{min}}}{4 \gamma}  + \Delta \right) \vec{u}(0)    \right]
\end{split}
\end{equation}
where $\vec{P}$ is a projection matrix that projects onto the subspace with the smallest eigenvalue $s_{\text{min}}$ (which may be one-dimensional), and where $\Delta := \sqrt{ \frac{s_{\text{min}}^2}{16 \gamma^2}    + \frac{1}{(c_* + \gamma)^2} } - \frac{1}{c_* + \gamma}$. Note that Eq. \ref{eq:align} implies that $\vec{W}$ aligns in the same direction as $\vec{u}$ and $\vec{v}$ (Fig. \ref{fig:align}c). Another interesting feature of this result is that $u_i(\tau)$ generally does not equal $v_i(\tau)$ even if they are initialized to the same value---noise breaks the `balancing' condition that can be assumed in the noise-free case \citep{domine_lazy_2025,bordelon_rnn_2025}. For large $s_i$, this discrepancy in scale can be quite large (Fig. \ref{fig:align}d).

\subsection{The outcome of alignment}

To very good approximation, we can assume $\vec{u}$ and $\vec{v}$ align and that $\vec{W}$ is a rank-one matrix supported on the same vector. Letting $\vec{u}_0$ denote that (normalized) vector,
\begin{equation}
\boxed{ \ \vec{u}(\tau) \approx u(\tau) \vec{u}_0 \hspace{0.5in} \vec{v}(\tau) \approx v(\tau) \vec{u}_0 \hspace{0.5in} \vec{W}(\tau) \approx [ \gamma - c(\tau) ] \vec{u}_0 \vec{u}_0^T \ }
\end{equation}
for some training-time-dependent coefficients $u(\tau)$, $v(\tau)$, and $c(\tau)$. Because gradient-flow dynamics does not change parameter directions under this ansatz, to characterize the next phase it suffices to analyze the dynamics of the scalars $u(\tau)$, $v(\tau)$, and $c(\tau)$. In the next few sections, we will use $\sigma^2$ to denote the noise level associated with the alignment direction (i.e., to denote the smallest eigenvalue of $\vec{S}$).

\section{Rich regime learning dynamics}

\begin{figure}[ht!]
\begin{center}
\includegraphics[width=0.8\textwidth]{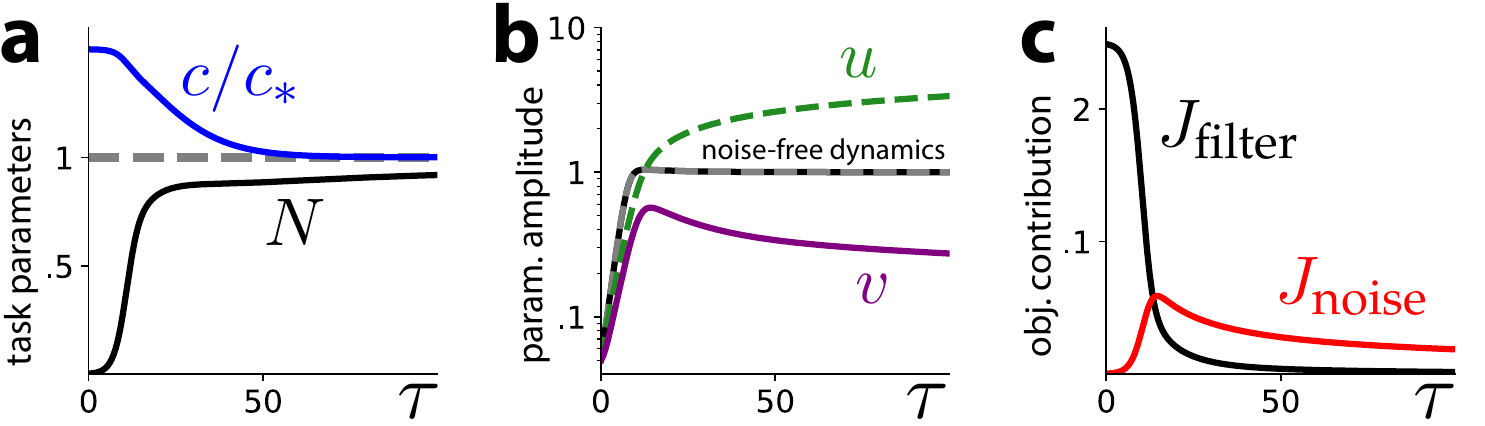}
\end{center}
\caption{\textbf{Rich regime learning dynamics.} For plotting, we take $c_* = 1$, $\sigma^2 = 1$, and $\kappa = 0$. (\textbf{a}) In the absence of regularization, learning dynamics drive $c$ toward its target, $c_*$, and the amplitude $N := u v$ toward $1$. (\textbf{b}) Unlike the noise-free ($\sigma^2 = 0$) dynamics, noisy dynamics produce asymmetric trajectories for $u$ and $v$, even if they are initialized to the same value. Asymptotically, $u v \approx 1$, $u$ continues to increase, and $v$ continues to decrease. (\textbf{c}) The filter-matching term and noise term are both driven to zero, although in this case the latter occurs on a slower time scale.}
\label{fig:rich}
\end{figure}

Alignment reduces an $N (N + 2)$-dimensional problem to a three-dimensional one. Post-alignment learning dynamics correspond to gradient descent on an effective objective (see Appendix \ref{app:low_dim_deriv})
\begin{equation} \label{eq:J_eff}
J =    \frac{1}{4 c(\tau)} \left[  u(\tau) v(\tau) - \frac{2 c(\tau)}{c_* + c(\tau)} \right]^2 + \left( \frac{1}{4 c_*} - \frac{c(\tau)}{[ c_* + c(\tau) ]^2} \right) +  \frac{\sigma^2 v(\tau)^2}{4 c(\tau)} + \frac{\kappa}{2} [ u(\tau)^2 + v(\tau)^2 ] 
\end{equation}
where we will refer to $\sigma^2 > 0$ as the ``noise level''. Learning dynamics in this regime follow 
\begin{equation} \label{eq:dyn_red}
\begin{split}
\frac{d}{d\tau} u(\tau) &=   \frac{v(\tau)}{c_* + c(\tau)} - \frac{u(\tau) v(\tau)^2}{2 c(\tau)} - \kappa u(\tau) \\
\frac{d}{d\tau} v(\tau) &=   \frac{u(\tau)}{c_* + c(\tau)} - \frac{v(\tau) [ \sigma^2  + u(\tau)^2 ]}{2 c(\tau)}  - \kappa v(\tau) \\
\frac{d}{d\tau} c(\tau) &=  \frac{v(\tau)^2 [ u(\tau)^2 + \sigma^2 ]}{4 c(\tau)^2}  -  \frac{u(\tau) v(\tau)}{[c_* + c(\tau) ]^2} \ .
\end{split}
\end{equation}
In the remainder of this section, we will assume no regularization ($\kappa = 0$). Let $N := u v$ denote the filter amplitude, so that $f(t) = N e^{- c t}$. $J$ is obviously minimized when $N = 1$ and $c = c_*$, since these choices set the first two terms to zero. Learning dynamics drive $N$ and $c$ toward those values (Fig. \ref{fig:rich}a). 

A key difference between the noisy and noise-free ($\sigma^2 = 0$) dynamics is that noise breaks the symmetry between $u$ and $v$. In the noise-free case, if $u$ and $v$ are initialized to the same value---i.e., if one makes a \textit{balancing} assumption \citep{domine_lazy_2025,bordelon_rnn_2025}---$u = v$ holds throughout training. In the noisy case, if there is no regularization, $u$ continues to grow while $v$ continues to shrink (Fig. \ref{fig:rich}b). In fact, Eq. \ref{eq:dyn_red} implies that
\begin{equation}
\frac{d}{d\tau} [ u(\tau)^2 - v(\tau)^2 ] = \frac{\sigma^2}{c(\tau)} v(\tau)^2 > 0  \ ,
\end{equation}
which means $u \to \infty$ and $v \to 0$. This makes sense, since taking $N = u v = 1$ and $v \to 0$ is precisely the combination that minimizes both the filter-matching term and the noise term (Fig. \ref{fig:rich}c).

\section{A noise-induced phase transition in rich regime dynamics}
\label{sec:ps1}

\begin{figure}[ht!]
\begin{center}
\includegraphics[width=\textwidth]{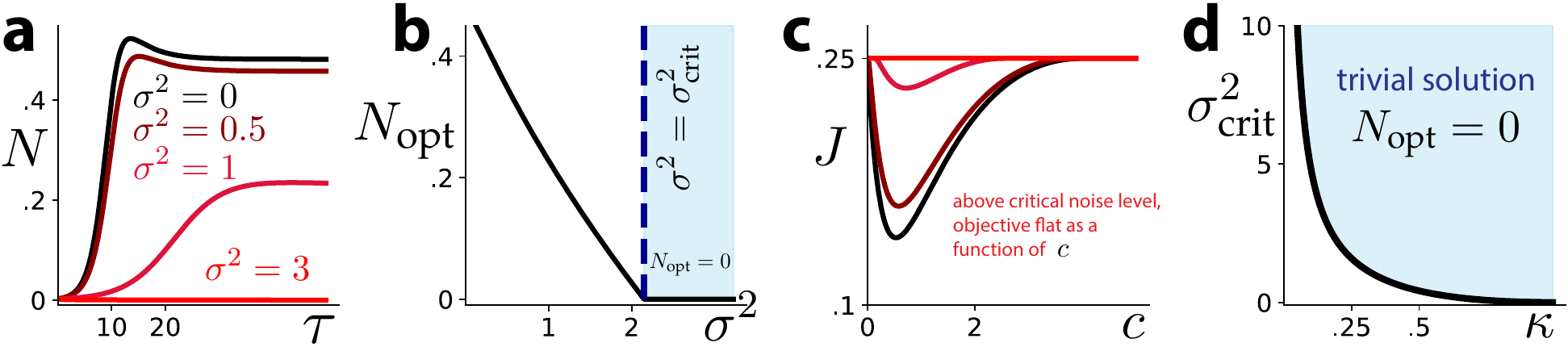}
\end{center}
\caption{\textbf{Excessive noise causes amplitude collapse.} For plotting, we take $c_* = 1$ and $\kappa = 0.2$. (\textbf{a}) Below a critical noise level $\sigma^2_{\text{crit}}$, learning dynamics drives the filter amplitude $N := u v$ toward a nontrivial fixed point. Above that noise level, $N$ is driven to zero. (\textbf{b}) The optimal amplitude $N_{\text{opt}}$ as a function of noise level. For $\sigma^2 \geq \sigma^2_{\text{crit}}$, the trivial solution with $f(t) = 0$ is optimal. (\textbf{c}) Below $\sigma^2_{\text{crit}}$, $J$ is minimized at a specific $c$ value. Above it, $J$ is insensitive to $c$ since $N = 0$. (\textbf{d}) The phase boundary $\sigma^2_{\text{crit}}(\kappa)$. The noise threshold separating trivial and nontrivial solutions depends on the regularization strength; it is higher when $\kappa$ is small, and $\sigma^2_{\text{crit}} \to \infty$ as $\kappa \to 0$.}
\label{fig:ps1}
\end{figure}

The behavior described in the previous section, aside from the asymmetry between $u$ and $v$ dynamics, is not substantially different from what \cite{bordelon_rnn_2025} observed in the noise-free case. However, the noisy case becomes significantly different when the regularization strength $\kappa$ is taken to be nonzero. Without regularization, we saw that the system minimizes noise by taking $u \to \infty$ and $v \to 0$; however, if high $u$ values are penalized, the system can become frustrated. If the noise level is sufficiently high, there is a \textit{phase transition}, and a trivial solution---which minimizes noise at the cost of a collapse in filter-matching performance---becomes optimal.

Below a critical noise level $\sigma^2_{\text{crit}}$ (which depends on the regularization strength $\kappa$), the filter amplitude $N := u v$ approaches a nonzero fixed point $N_{\text{opt}}$. When $\kappa = 0$, this equals one, but it decreases as $\kappa$ increases, since the scales of $u$ and $v$ are penalized by the regularization term. Beyond the critical noise level, dynamics drive $N$ to zero (Fig. \ref{fig:ps1}a), and moreover $J$ is minimized at $N_{\text{opt}} = 0$ (Fig. \ref{fig:ps1}b). 

One can show algebraically (see Appendix \ref{app:ps1}) that, after minimizing $J$ over $u$ and $v$, it becomes
\begin{equation}
\begin{split}
J(c) &= \frac{1}{4 c_*} - c \left[  \frac{1}{c_* + c} - \kappa  \sqrt{1 + \frac{\sigma^2}{2 \kappa c}}  \right]_+^2 \ .
\end{split}
\end{equation}
where $[ ... ]_+$ denotes rectification (i.e., a ReLU). This simplified objective only has a nontrivial solution if the bracketed quantity is greater than zero; otherwise, $J$ becomes independent of $c$ (Fig. \ref{fig:ps1}c). 

We can identify the corresponding phase boundary $\sigma^2_{\text{crit}}(\kappa)$ by determining the noise level at which the bracketed quantity and its derivative equal zero. We find (see Appendix \ref{app:ps1}) that
\begin{equation}
\boxed{ \ \sigma^2_{\text{crit}}(\kappa) =   \frac{[ 1 - q(\kappa)^2 ]^2}{q(\kappa)} \ }
\end{equation}
for $\kappa < 1/c_*$, and $\sigma^2_{\text{crit}} = 0$ for $\kappa \geq 1/c_*$. Here,
\begin{equation}
q(\kappa) := \frac{2}{\sqrt{3}} \sinh\left[ \frac{1}{3} \text{arsinh}(3 \sqrt{3} \kappa c_* ) \right] \hspace{0.3in} \text{and} \hspace{0.3in} \text{arsinh}(x) := \log( x + \sqrt{x^2 + 1} ) \ .
\end{equation}
The phase boundary is plotted in Fig. \ref{fig:ps1}d. It has a few features worth emphasizing. First, since $q(\kappa) \to 0$ as $\kappa \to 0$, $\sigma^2_{\text{crit}}$ diverges to infinity in this limit; this is consistent with the observation that there is no phase transition without regularization. Second, for small $\kappa$, we have the comparatively simple expressions
\begin{equation}
q(\kappa) \approx 2 c_* \kappa \hspace{1in} \sigma^2_{\text{crit}}(\kappa) \approx \frac{1}{2 c_* \kappa}
\end{equation}
which qualitatively match Fig. \ref{fig:ps1}d. In summary, if there is too much noise, it is almost as if the system's learning dynamics seek an `escape route': parameters are pushed into trivial regimes (here, the filter amplitude $N$ is taken to zero) in order to minimize noise at the expense of task accuracy.

\section{A second phase transition when scale changes are impossible}
\label{sec:ps2}

\begin{figure}[ht!]
\begin{center}
\includegraphics[width=\textwidth]{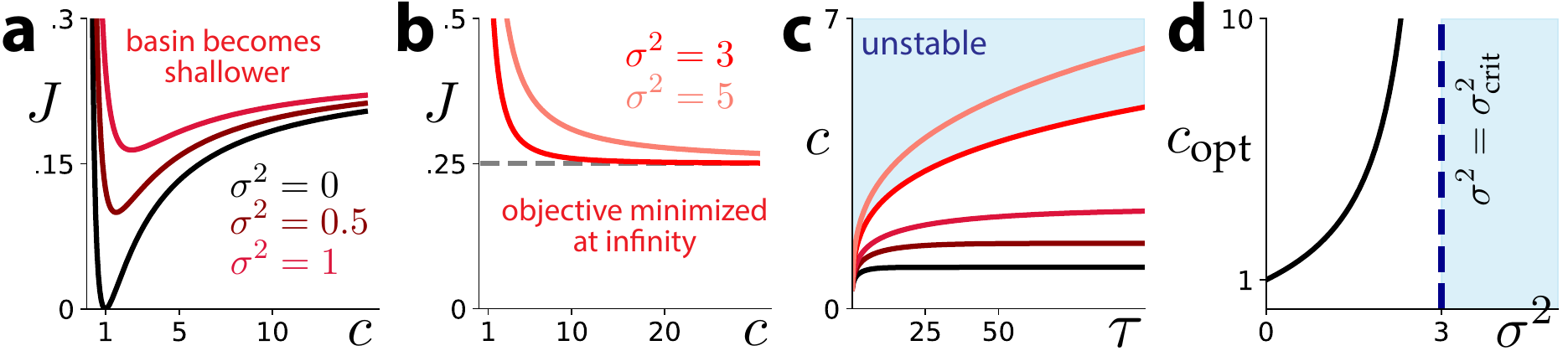}
\end{center}
\caption{\textbf{Excessive noise produces an infinitely fast filter.} $c_* = 1$ assumed throughout. (\textbf{a}) When the read-in and readout scales are fixed, $J$ is a function of $c$ only, and for low enough noise it has a single attractor basin at $c = c_{\text{opt}}$. As $\sigma^2$ increases, this basin becomes increasingly shallow. (\textbf{b}) For $\sigma^2 \geq \sigma^2_{\text{crit}} = 3$, $J$ is a decreasing function of $c$ with no finite minimum. (\textbf{c}) Below the noise threshold, dynamics drive $c$ toward the unique stable fixed point; above it, $c$ grows without bound. (\textbf{d}) $c_{\text{opt}}$ diverges like $1/(\sigma^2_{\text{crit}} - \sigma^2)$ just below the noise threshold.}
\label{fig:ps2}
\end{figure}

We just saw how, for sufficiently large intrinsic noise $\sigma^2 > 0$, the system may converge to a trivial solution with zero amplitude, i.e., $u v = 0$. One may wonder whether this behavior is peculiar to our setup---for example, whether it depends on trainable read-in and readout weights. Interestingly, if $u$ and $v$ are frozen, a different type of noise-induced phase transition is possible. 

Consider the $u = v = 1$ specialization of the simplified objective (Eq. \ref{eq:J_eff}):
\begin{equation}
J =    \frac{1}{4 c(\tau)} \left[  1 - \frac{2 c(\tau)}{c_* + c(\tau)} \right]^2 + \frac{1}{4 c_*} - \frac{c(\tau)}{[ c_* + c(\tau) ]^2} +  \frac{\sigma^2}{4 c(\tau)}  = \frac{1 + \sigma^2}{4 c(\tau)} - \frac{1}{c_* + c(\tau)} + \frac{1}{4 c_*} 
\end{equation}
where we have removed the regularization term because it is constant. The corresponding learning dynamics are (see Appendix \ref{app:ps2})
\begin{equation} \label{eq:dyn_ps2}
\frac{d}{d\tau} c = \frac{[ \sqrt{1 + \sigma^2} (c_* + c) + 2 c] }{4 c^2 (c_* + c)^2} [ \sqrt{1 + \sigma^2} (c_* + c) - 2 c]
\end{equation}
For sufficiently small noise, these dynamics have a stable positive fixed point at
\begin{equation}
c_{\text{opt}} = \frac{\sqrt{1 + \sigma^2} \ c_*}{2 - \sqrt{1 + \sigma^2}} \ .
\end{equation}
As one might expect, $c_{\text{opt}}$ equals the target rate $c_*$ when there is no noise. However, as $\sigma^2$ increases, the corresponding basin of $J$ becomes increasingly shallow (Fig. \ref{fig:ps2}a). Pushing the noise level higher than $\sigma^2_{\text{crit}} = 3$ causes this basin to disappear (Fig. \ref{fig:ps2}b), and for $J$ to be minimized at $c = \infty$. Indeed, Eq. \ref{eq:dyn_ps2} has $dc/d\tau > 0$ for all $c$ if $\sigma^2 \geq 3$, so learning drives $c$ toward infinity (Fig. \ref{fig:ps2}c). Just below $\sigma^2_{\text{crit}}$, the optimal $c$ value diverges as $1/(\sigma^2_{\text{crit}} - \sigma^2)$ (Fig. \ref{fig:ps2}d). We conclude that, although the precise representational `escape route' may vary, the system may be driven toward trivial solutions that minimize noise when $\sigma^2$ is excessively large.

\section{Discussion}

We developed a theory of noisy linear RNN learning dynamics and mathematically showed that noise can decisively shape representation learning. Consistent with \cite{bordelon_rnn_2025}, in the `rich' regime, where parameters are initialized to small values, a low-rank task (here, matching a single leaky integration time scale) yields an approximately low-rank solution. But our findings differ from those of \cite{bordelon_rnn_2025} in three important ways. First, \textit{noise strongly modulates the initial alignment phase}, and $\vec{u}$, $\vec{v}$, and $\vec{W}$ preferentially align along the least noisy eigenvector (or eigenvectors) of $\vec{S}$, rather than simply in the average direction of the initial $\vec{u}$ and $\vec{v}$; this is interesting in part because it makes the network's solution depend much less on its initial condition. 

Second, \textit{noise produces asymmetry between the read-in and readout weights}, and in the second phase $u$ tends to become large while $v$ tends to become small. This is true even if $u$ and $v$ are initialized to identical values because noise breaks the \textit{balancing} property that holds in the noise-free case. Finally, \textit{noise produces multiple types of phase transitions}, because minimizing noise can compete with achieving good task performance. If $\vec{u}$ and $\vec{v}$ are trainable, we find that excessive noise drives the network toward a solution whose filter amplitude $N := u v$ equals zero; interestingly, this only occurs if regularization constrains the scales of $u$ and $v$. If only $\vec{W}$ is trainable, excessive noise drives the system to $f(t) = 0$ in a different way, by making the filter's frequency infinitely fast.

In other machine learning settings, noise is known to shape what types of representations systems learn. Noise in optimization due to using stochastic rather than full-batch gradient descent tends to produce convergence to broader loss landscape minima that generalize better \citep{le_sgd_2018,smith_sgd_2020}. Noise in diffusion generative modeling may similarly improve generalization, albeit potentially through a different mechanism \citep{vastola_generalization_2025}. Our work suggests that noise shapes RNN representation learning, but does not necessarily produce \textit{better} solutions---only \textit{different} ones. Here, noise reduces solution-space degeneracy by constraining the direction of weight alignment and the relative scales of the read-in and readout weights. If there is too much noise, it is clear that noise has deleterious effects, since it drives the system toward trivial solutions.

\paragraph{Limitations and future work.} For tractability's sake, we made a large number of simplifying assumptions: we assumed a \textit{whitened} scalar input; a \textit{linear} network; a simple filter-matching task involving learning \textit{one} time scale; and near-zero parameter initializations that yield low-rank learning dynamics. Although the work we build on makes similar assumptions, it may be possible to relax some of them. For example, it may be possible to study nonlinear networks by linearizing them around a fixed point (either in state space or parameter space), or by exploiting dynamical mean-field theory machinery \citep{bordelon_self-consistent_2023}. One class of tasks which is more realistic, but whose learning dynamics may remain tractable, is target pursuit \citep{vastola2026solvable}.



\subsubsection*{Acknowledgments}

KR was funded by the NIH (RF1DA056403, U01NS136507), James S. McDonnell Foundation (220020466), Simons Foundation (Pilot Extension-00003332-02), McKnight Endowment Fund, CIFAR Azrieli Global Scholar Program, NSF (2046583), a Harvard Medical School Neurobiology Lefler Small Grant Award, and a Harvard Medical School Dean’s Innovation Award. This work has been made possible in part by a gift from the Chan Zuckerberg Initiative Foundation to establish the Kempner Institute for the Study of Natural and Artificial Intelligence at Harvard University.

\subsection*{AI use statement}

In this work, we used generative AI tools for brainstorming, literature search, to assist with checking the mathematical correctness of our results, and to assist with code generation. The authors have checked all text and results and take full responsibility for them.

%

\bibliography{rnn_bib}
\bibliographystyle{iclr2027_conference}

\clearpage

\appendix

\section{Objective function and derivation of learning dynamics}
\label{app:obj}

In this appendix, we simplify our original objective function and derive expressions for the associated gradient descent learning dynamics (in the gradient-flow limit that yields ODEs rather than a discrete-time recurrence). 

\subsection{Setup}

Recall from Sec. \ref{sec:math} that the objective function has the form
\begin{equation}
\boxed{ \ J := \lim_{T \to \infty} \frac{1}{T} \int_0^T \frac{1}{2} \mathbb{E}_{x, \vec{r}}\left\{ \ [ y(t) - y_*(t) ]^2 \ \right\} \ dt + \frac{\kappa}{2} \Vert \vec{u} \Vert^2  + \frac{\kappa}{2} \Vert \vec{v} \Vert^2   \ }
\end{equation}
where $y(t) = \vec{v}^T \vec{r}(t)$ is the RNN's output and
\begin{equation}
y_*(t) := \int_0^t e^{- c_* t'} x(t - t') \ dt'  = \int_0^t f_*(t') x(t - t') \ dt'
\end{equation}
is the target output. If we assume zero initial activity (i.e., $\vec{r}(0) = \vec{0}$), then as we mention in Sec. \ref{sec:math}, 
\begin{equation}
p(\vec{r} , t | \vec{r}_0 = \vec{0}, 0) = \mathcal{N}\left(  \int_0^t e^{\vec{A} t'} \vec{u} \ x(t-t') \ dt'    ,   \int_0^t  e^{\vec{A} t'} \vec{S} e^{\vec{A}^T t'} \ dt'  \right)  \ .
\end{equation}
This implies that the output $y(t)$ has statistics
\begin{equation}
\begin{split}
\mathbb{E}_{\vec{r}}[ y(t) ] &= \int_0^t \vec{v}^T e^{\vec{A} t'} \vec{u} \ x(t-t') \ dt' = \int_0^t f(t') \ x(t-t') \ dt' \\
\text{var}_{\vec{r}}[ y(t) ] &= \vec{v}^T \vec{\Sigma} \vec{v}
\end{split}
\end{equation}
where $f(t) := e^{- \gamma t} \vec{v}^T e^{\vec{W} t} \vec{u}$ is the learned filter and the covariance $\vec{\Sigma}$ is defined as
\begin{equation}
\vec{\Sigma} :=  \int_0^t  e^{\vec{A} t'} \vec{S} e^{\vec{A}^T t'} \ dt' \ .
\end{equation}
Also recall that the input signal $x(t)$ is assumed to have white noise statistics, i.e.,  $\mathbb{E}_x[ x(t) ] = 0$ and $\mathbb{E}_x[ x(t) x(t - t') ] = \delta(t - t')$.

\subsection{Simplifying the objective}

\paragraph{Decomposing into a bias and variance term.} First, average over $\vec{r}$:
\begin{equation}
\begin{split}
J &= \lim_{T \to \infty} \frac{1}{T} \int_0^T \frac{1}{2} \mathbb{E}_x\left\{ \   [ \mathbb{E}_{\vec{r}}[ y(t) ] - y_*(t) ]^2 + \text{var}_{\vec{r}}[ y(t) ]    \ \right\} \ dt + \frac{\kappa}{2} \Vert \vec{u} \Vert^2  + \frac{\kappa}{2} \Vert \vec{v} \Vert^2   \\
&= \lim_{T \to \infty} \frac{1}{T} \int_0^T \frac{1}{2} \mathbb{E}_x\left\{ \   [ \mathbb{E}_{\vec{r}}[ y(t) ] - y_*(t) ]^2 \ \right\}  + \frac{1}{2} \text{var}_{\vec{r}}[ y(t) ]    \ dt + \frac{\kappa}{2} \Vert \vec{u} \Vert^2  + \frac{\kappa}{2} \Vert \vec{v} \Vert^2  \\
&= \lim_{T \to \infty} \frac{1}{T} \int_0^T \frac{1}{2} \mathbb{E}_x\left\{ \   [ \mathbb{E}_{\vec{r}}[ y(t) ] - y_*(t) ]^2 \ \right\}  + \frac{1}{2} \vec{v}^T \vec{\Sigma}(t) \vec{v}   \ dt  + \frac{\kappa}{2} \Vert \vec{u} \Vert^2  + \frac{\kappa}{2} \Vert \vec{v} \Vert^2  
\end{split}
\end{equation}
where the second equality uses the fact that variance in $y(t)$ is input-independent.

\paragraph{Simplifying the mean term.} The second term is input-independent; it represents noise due to recurrent dynamics. The first term can be written in terms of filters:
\begin{equation}
\begin{split}
\ &= \lim_{T \to \infty} \frac{1}{T} \int_0^T \frac{1}{2} \mathbb{E}_{x}\left\{ \ \int_0^t \int_0^t [ f(s) - f_*(s) ] [ f(s') - f_*(s') ] \ x(t-s) x(t - s') \ \right\} \ ds ds' dt \\
&= \lim_{T \to \infty} \frac{1}{T} \int_0^T \frac{1}{2}  \ \int_0^t \int_0^t   [ f(s) - f_*(s) ] [ f(s') - f_*(s') ]  \ \mathbb{E}_x[x(t-s) x(t - s')] \ ds ds' dt \\
&= \lim_{T \to \infty} \frac{1}{T} \int_0^T \frac{1}{2}  \ \int_0^t \int_0^t  [ f(s) - f_*(s) ] [ f(s') - f_*(s') ]  \ \delta(s - s') \ ds ds' dt \\
&= \lim_{T \to \infty} \frac{1}{T} \int_0^T \frac{1}{2}  \ \int_0^t   [ f(s) - f_*(s) ]^2  \ ds  dt \ .
\end{split}
\end{equation}
This can be simplified with a clever rearrangement. Note,
\begin{equation}
\begin{split}
 \lim_{T \to \infty} \frac{1}{2 T} \int_0^T \int_0^t   [ f(s) - f_*(s) ]^2 \ ds  dt &=  \lim_{T \to \infty} \frac{1}{2 T} \int_0^T \int_{s}^T   [ f(s) - f_*(s) ]^2 \ dt  ds \\
 &=  \lim_{T \to \infty} \frac{1}{2} \int_0^T \left( 1 - \frac{s}{T} \right)   [ f(s) - f_*(s) ]^2 \  ds \ .
\end{split}
\end{equation}
If the eigenvalues of $\vec{A}$ have negative real parts (i.e., if dynamics is stable), then we can drop the $s/T$ term, since it vanishes in the $T \to \infty$ limit. We conclude that
\begin{equation}
 J_{\text{filter}} =   \frac{1}{2} \int_0^{\infty}  [ f(t) - f_*(t) ]^2 \  dt \ .
\end{equation}

\paragraph{Simplifying the variance term.} The noise term can be rewritten as
\begin{equation}
\begin{split}
& \ \lim_{T \to \infty} \frac{1}{2 T} \int_0^T \vec{v}^T \vec{\Sigma}(t) \vec{v} \ dt \\
=& \ \vec{v}^T \left\{ \ \lim_{T \to \infty} \frac{1}{2 T} \int_0^T  \int_0^t  e^{\vec{A} s} \vec{S} e^{\vec{A}^T s} \ ds  dt \ \right\} \vec{v} \\
=& \ \vec{v}^T \left\{ \ \lim_{T \to \infty} \frac{1}{2 T} \int_0^T  \int_{s}^T  e^{\vec{A} s} \vec{S} e^{\vec{A}^T s} \ dt  ds \ \right\} \vec{v} \\
=& \ \vec{v}^T \left\{ \ \lim_{T \to \infty} \frac{1}{2} \int_0^T    \left( 1 - \frac{s}{T} \right) e^{\vec{A} s} \vec{S} e^{\vec{A}^T s} \   ds \ \right\} \vec{v} \ .
\end{split}
\end{equation}
Once again, if dynamics are stable, we can drop the $s/T$ term in the infinite $T$ limit. We obtain
\begin{equation}
J_{\text{noise}} = \frac{1}{2} \int_0^{\infty}   \vec{v}^T e^{\vec{A} t} \vec{S} e^{\vec{A}^T t} \vec{v}  \   dt \ .
\end{equation}

\paragraph{Total objective.}

Adding all of the terms, we obtain the simplified objective
\begin{equation}
\boxed{ \ 
J(\vec{u}, \vec{v}, \vec{W}) =  \int_0^{\infty}    \frac{1}{2} \left[ f(t) - f_*(t) \right]^2 +  \frac{e^{- 2 \gamma t}}{2}  \vec{v}^T e^{\vec{W} t} \vec{S} e^{\vec{W}^T t} \vec{v} \   dt   + \frac{\kappa}{2} \Vert \vec{u} \Vert^2  + \frac{\kappa}{2} \Vert \vec{v} \Vert^2 \ } \ .
\end{equation}

\subsection{Useful derivatives}

A few derivatives are necessary to compute the gradients of $J$. First, the derivative of the matrix exponential  with respect to $\vec{W}$ has components
\begin{equation}
\frac{\partial}{\partial W_{i j}} e^{\vec{W} t} = \int_0^t dt' \ e^{\vec{W} t'} \vec{e}_i \vec{e}_j^T e^{\vec{W} (t - t')} \ .
\end{equation}
Similarly, the derivative of its transpose is
\begin{equation}
\frac{\partial}{\partial W_{i j}} e^{\vec{W}^T t} = \int_0^t dt' \ e^{\vec{W}^T t'} \vec{e}_j \vec{e}_i^T e^{\vec{W}^T (t - t')} \ .
\end{equation}
The derivative of the learned filter $f(t)$ with respect to $\vec{W}$ has components
\begin{equation}
\frac{\partial}{\partial W_{i j}} f(t) = e^{- \gamma t} \int_0^t dt' \ \left[ e^{\vec{W}^T t'} \vec{v} \right]_i  \left[ e^{\vec{W} (t - t')} \vec{u} \right]_j = e^{- \gamma t} \int_0^t dt' \ e^{\vec{W}^T t'} \vec{v} \vec{u}^T e^{\vec{W}^T (t - t')} \ .
\end{equation}
Finally, the derivative of the variance term with respect to $\vec{W}$ has components
\begin{equation}
\begin{split}
& \ \frac{\partial}{\partial W_{i j}}  \vec{v}^T e^{ \vec{W} t} \vec{S} e^{ \vec{W}^T t} \vec{v} \\
=& \ \int_0^t dt' \ \vec{v}^T e^{\vec{W} t'} \vec{e}_i \vec{e}_j^T e^{\vec{W} (t - t')} \vec{S} e^{\vec{W}^T t} \vec{v} +  \int_0^t dt' \ \vec{v}^T e^{\vec{W} t} \vec{S} e^{\vec{W}^T t'} \vec{e}_j \vec{e}_i^T e^{\vec{W}^T (t - t')} \vec{v}  \\
=& \ \int_0^t dt' \  e^{\vec{W}^T t'} \vec{v} \vec{v}^T e^{\vec{W} t} \vec{S} e^{\vec{W}^T (t - t')} + \int_0^t dt' \ e^{\vec{W}^T (t - t')} \vec{v} \vec{v}^T e^{\vec{W} t} \vec{S} e^{\vec{W}^T t'}   \ .
\end{split}
\end{equation}

\subsection{Derivation of learning dynamics}

Using the aforementioned derivatives, we can immediately compute the relevant gradients of $J$:
\begin{equation}
\begin{split}
\frac{\partial J}{\partial \vec{u}} &= \int_0^{\infty} [ f(t) - f_*(t) ] \frac{\partial f(t)}{\partial \vec{u}} dt + \kappa \vec{u} \\
&=  \int_0^{\infty} [ f(t) - f_*(t) ] e^{- \gamma t} e^{\vec{W}^T t} \vec{v}  \ dt + \kappa \vec{u} \\
\frac{\partial J}{\partial \vec{v}} &= \int_0^{\infty} [ f(t) - f_*(t) ] \frac{\partial f(t)}{\partial \vec{v}} +  e^{- 2 \gamma t} e^{ \vec{W} t} \vec{S} e^{ \vec{W}^T t} \vec{v}  \  dt + \kappa \vec{v} \\
&= \int_0^{\infty} [ f(t) - f_*(t) ] e^{- \gamma t} e^{\vec{W} t} \vec{u}  +  e^{- 2 \gamma t} e^{ \vec{W} t} \vec{S} e^{ \vec{W}^T t} \vec{v}  \  dt + \kappa \vec{v} \\
\frac{\partial J}{\partial \vec{W}} &= \int_0^{\infty} [ f(t) - f_*(t) ] e^{- \gamma t} \int_0^t  \ e^{\vec{W}^T t'} \vec{v} \vec{u}^T e^{\vec{W}^T (t - t')} \ dt' dt \\
&+ \int_0^{\infty} \frac{e^{-2\gamma t}}{2} \int_0^t   e^{\vec{W}^T t'} \vec{v} \vec{v}^T e^{\vec{W} t} \vec{S} e^{\vec{W}^T (t - t')} \ dt' dt + \int_0^{\infty} \frac{e^{-2\gamma t}}{2} \int_0^t  \ e^{\vec{W}^T (t - t')} \vec{v} \vec{v}^T e^{\vec{W} t} \vec{S} e^{\vec{W}^T t'}  \ dt' dt \ .
\end{split}
\end{equation}
This result immediately implies that our gradient descent learning dynamics have the form
\begin{equation}
\begin{split}
\frac{d}{d\tau} \vec{u} &=  \int_0^{\infty} [ f_*(t) - f(t) ] e^{- \gamma t} e^{\vec{W}^T t} \vec{v}  \ dt - \kappa \vec{u}  \\
\frac{d}{d\tau} \vec{v} &= \int_0^{\infty} [ f_*(t) - f(t) ] e^{- \gamma t} e^{\vec{W} t} \vec{u}  -  e^{- 2 \gamma t} e^{ \vec{W} t} \vec{S} e^{ \vec{W}^T t} \vec{v}  \  dt - \kappa \vec{v} \\
\frac{d}{d\tau} \vec{W} &= \int_0^{\infty} [ f_*(t) - f(t) ] e^{- \gamma t} \int_0^t  e^{\vec{W}^T t'} \vec{v} \vec{u}^T e^{\vec{W}^T (t - t')} \ dt' dt\\
&- \int_0^{\infty} \frac{e^{-2\gamma t}}{2} \int_0^t  \  e^{\vec{W}^T t'} \vec{v} \vec{v}^T e^{\vec{W} t} \vec{S} e^{\vec{W}^T (t - t')} \ dt' dt - \int_0^{\infty} \frac{e^{-2\gamma t}}{2} \int_0^t  \ e^{\vec{W}^T (t - t')} \vec{v} \vec{v}^T e^{\vec{W} t} \vec{S} e^{\vec{W}^T t'}  \ dt' dt \ .
\end{split}
\end{equation}
By a change of variables $t' \to t - t'$, one can show that the two noise terms in the $\vec{W}$ dynamics are identical. This allows the final result to be written in a slightly more compact form:
\begin{equation}
\begin{split}
\frac{d}{d\tau} \vec{u} &=  \int_0^{\infty} [ f_*(t) - f(t) ] e^{- \gamma t} e^{\vec{W}^T t} \vec{v}  \ dt - \kappa \vec{u}  \\
\frac{d}{d\tau} \vec{v} &= \int_0^{\infty} [ f_*(t) - f(t) ] e^{- \gamma t} e^{\vec{W} t} \vec{u}  -  e^{- 2 \gamma t} e^{ \vec{W} t} \vec{S} e^{ \vec{W}^T t} \vec{v}  \  dt - \kappa \vec{v} \\
\frac{d}{d\tau} \vec{W} &= \int_0^{\infty} [ f_*(t) - f(t) ] e^{- \gamma t} \int_0^t  e^{\vec{W}^T t'} \vec{v} \vec{u}^T e^{\vec{W}^T (t - t')} \ dt' dt\\
&- \int_0^{\infty} e^{-2\gamma t} \int_0^t \  e^{\vec{W}^T t'} \vec{v} \vec{v}^T e^{\vec{W} t} \vec{S} e^{\vec{W}^T (t - t')} \ dt' dt  \ .
\end{split}
\end{equation}

\subsection{Learning dynamics given small initial weights}

We can expand these learning dynamics to leading order in $\vec{u}$, $\vec{v}$, and $\vec{W}$ (supposing they are all of order $\epsilon > 0$, say). Doing so, we find that
\begin{equation}
\begin{split}
\frac{d}{d\tau} \vec{u} &\approx  \int_0^{\infty}  f_*(t) e^{- \gamma t} \vec{v}  \ dt - \kappa \vec{u}  \\
\frac{d}{d\tau} \vec{v} &\approx \int_0^{\infty}  f_*(t) e^{- \gamma t}  \vec{u}  -  e^{- 2 \gamma t}  \vec{S} \vec{v}  \  dt - \kappa \vec{v} \\
\frac{d}{d\tau} \vec{W} &\approx \int_0^{\infty} f_*(t)  e^{- \gamma t} \int_0^t \ \vec{v} \vec{u}^T  \ dt' dt - \int_0^{\infty} e^{-2\gamma t} \int_0^t \  \vec{v} \vec{v}^T \vec{S}  \ dt' dt \ .
\end{split}
\end{equation}
We can be more explicit using the fact that $f_*(t) = e^{- c_* t}$. Simplifying yields
\begin{equation}
\begin{split}
\frac{d}{d\tau} \vec{u} &\approx  \frac{1}{c_* + \gamma} \vec{v} - \kappa \vec{u}   \\
\frac{d}{d\tau} \vec{v} &\approx \frac{1}{c_* + \gamma}  \vec{u}  -  \frac{1}{2 \gamma} \vec{S} \vec{v}  - \kappa \vec{v} \\
\frac{d}{d\tau} \vec{W} &\approx \int_0^{\infty} f_*(t)  e^{- \gamma t} \int_0^t \ \vec{v} \vec{u}^T  \ dt' dt - \int_0^{\infty} e^{-2\gamma t} \int_0^t  \  \vec{v} \vec{v}^T \vec{S}  \ dt' dt  \\
&\approx \frac{1}{(c_* + \gamma)^2} \vec{v} \vec{u}^T  - \frac{1}{4 \gamma^2}  \vec{v} \vec{v}^T \vec{S}  \ .
\end{split}
\end{equation}

\clearpage

\section{Reservoir computing case}
\label{app:reservoir}

In this appendix, we work out the learning dynamics of the \textit{reservoir computing} model in which only the readout $\vec{v}$ is trained. Its primary virtue is that it illustrates, even when learning dynamics are linear, how noise produces representational pressure. For simplicity, we ignore regularization and set $\kappa = 0$.

Recall from Appendix \ref{app:obj} that $\vec{v}$ has dynamics
\begin{equation}
\begin{split}
\frac{d}{d\tau} \vec{v} &= \int_0^{\infty} [ f_*(t) - f(t) ] e^{- \gamma t} e^{\vec{W} t} \vec{u}  -  e^{- 2 \gamma t} e^{ \vec{W} t} \vec{S} e^{ \vec{W}^T t} \vec{v}  \  dt \\
&= \int_0^{\infty}  e^{- (c_* + \gamma) t} e^{\vec{W} t} \vec{u} - e^{- 2\gamma t} e^{\vec{W} t} \vec{u} \vec{u}^T e^{\vec{W}^T t} \vec{v}  -  e^{- 2 \gamma t} e^{ \vec{W} t} \vec{S} e^{ \vec{W}^T t} \vec{v}  \  dt \\
&= \left[ \int_0^{\infty}  e^{- (c_* + \gamma) t} e^{\vec{W} t} \ dt \right] \vec{u}  - \left[  \int_0^{\infty}  e^{- 2\gamma t} e^{\vec{W} t} (\vec{u} \vec{u}^T + \vec{S}) e^{\vec{W}^T t} \ dt \right] \vec{v}   \ .
\end{split}
\end{equation}
These dynamics are linear, and the matrix that controls which directions of $\vec{v}$ are learned first is
\begin{equation}
\vec{M} :=   \int_0^{\infty}  e^{- 2\gamma t} e^{\vec{W} t} (\vec{u} \vec{u}^T + \vec{S}) e^{\vec{W}^T t} \ dt \ .
\end{equation}
The fixed point of these dynamics is
\begin{equation}
\boxed{ \ \vec{v}_* = \left(  \int_0^{\infty}  e^{- 2\gamma t} e^{\vec{W} t} (\vec{u} \vec{u}^T + \vec{S}) e^{\vec{W}^T t} \ dt \right)^{-1} \left( \int_0^{\infty}  e^{- (c_* + \gamma) t} e^{\vec{W} t} \ dt \right) \vec{u} \ } \ .
\end{equation}

\paragraph{Informative simplification.} Assume for simplicity that $\vec{S}$ is diagonal and $\vec{W} = (\gamma - c_*) \vec{I}$, i.e., that all state-space directions are compatible with the target filter. In this case,
\begin{equation}
\begin{split}
\int_0^{\infty}  e^{- (c_* + \gamma) t} e^{\vec{W} t} \ dt &= \frac{1}{2 c_*} \\
\int_0^{\infty}  e^{- 2\gamma t} e^{\vec{W} t} (\vec{u} \vec{u}^T + \vec{S}) e^{\vec{W}^T t} \ dt &= \frac{1}{2 c_*} (\vec{u} \vec{u}^T + \vec{S})
\end{split}
\end{equation}
so the optimal readout is
\begin{equation}
\boxed{ \ \vec{v}_* = (\vec{u} \vec{u}^T + \vec{S})^{-1} \vec{u} = \left( \vec{S}^{-1} - \frac{\vec{S}^{-1} \vec{u} \vec{u}^T \vec{S}^{-1}}{1 + \vec{u}^T \vec{S}^{-1} \vec{u}} \right) \vec{u} = \frac{\vec{S}^{-1} \vec{u}}{1 + \vec{u}^T \vec{S}^{-1} \vec{u}} \ }
\end{equation}
For a large read-in vector $\vec{u}$, this approximately equals
\begin{equation}
\vec{v}_* \approx \frac{\vec{S}^{-1} \vec{u}}{\vec{u}^T \vec{S}^{-1} \vec{u}}
\end{equation}
which weighs each input direction by how noisy it is. This prescription also yields the correct optimal filter, since
\begin{equation}
\vec{u}^T \vec{v}_* \approx \frac{\vec{u}^T \vec{S}^{-1} \vec{u}}{\vec{u}^T \vec{S}^{-1} \vec{u}} = 1 \ ,
\end{equation}
which implies
\begin{equation}
f(t) = \vec{v}^T_* e^{\vec{A} t} \vec{u} =  e^{- c_* t} \vec{v}_*^T \vec{u} \approx e^{- c_* t} \ .
\end{equation}
This approximate solution is intuitive since it is precisely the choice of $\vec{v}$ that, for this choice of $\vec{W}$ and $\vec{u}$, minimizes the quadratic noise term subject to the constraint that $\vec{u}^T \vec{v} = 1$.

\clearpage

\section{Alignment dynamics}
\label{app:align}

In this appendix, we solve the dynamics associated with the alignment phase (see Sec. \ref{sec:align}). Using the small-parameter gradient descent updates from Appendix \ref{app:obj}, we have
\begin{equation} \label{eq:align_app}
\frac{d}{d\tau} \vec{u} \approx  \frac{\vec{v} }{c_* + \gamma} - \kappa \vec{u} \hspace{0.28in} \frac{d}{d\tau} \vec{v} \approx \frac{\vec{u}}{c_* + \gamma}    -  \frac{\vec{S} \vec{v}}{2 \gamma} - \kappa \vec{v} \hspace{0.28in}  \frac{d}{d\tau} \vec{W} \approx \frac{\vec{v} \vec{u}^T }{(c_* + \gamma)^2}  - \frac{ \vec{v} \vec{v}^T \vec{S} }{4 \gamma^2}  \ .
\end{equation}
Because $\vec{u}$ and $\vec{v}$ dynamics form a closed system, we will begin by studying $\vec{u}$ and $\vec{v}$ only. Once we determine how $\vec{u}$ and $\vec{v}$ change with time, we will use them to determine how $\vec{W}$ changes.

\subsection{Solving read-in and readout dynamics}

Naively, $\vec{u}$ and $\vec{v}$ together determine a $2N$-dimensional system of ODEs, since
\begin{equation}
\begin{split}
\frac{d}{d\tau} \vec{u} &= - \kappa \vec{u} + \frac{\vec{v} }{c_* + \gamma}  \\
\frac{d}{d\tau} \vec{v} &= \frac{\vec{u}}{c_* + \gamma}    -  \left(  \kappa + \frac{\vec{S}}{2 \gamma}  \right) \vec{v} \ .
\end{split}
\end{equation}
Note that, to ease notation, we will write equalities instead of $\approx$. 

This system substantially simplifies if we work in a basis of eigenvectors of $\vec{S}$. Along the $i$-th such eigenvector, if $s_i > 0$ denotes the corresponding eigenvalue, 
\begin{equation}
\begin{split}
\frac{d}{d\tau} u_i &= - \kappa u_i + \frac{v_i}{c_* + \gamma}  \\
\frac{d}{d\tau} v_i &= \frac{u_i}{c_* + \gamma}    - \left( \kappa + \frac{s_i}{2 \gamma} \right) v_i \ .
\end{split}
\end{equation}
\paragraph{Relevant eigenvalues and eigenvectors.} This is just a two-dimensional system governed by the $2 \times 2$ matrix
\begin{equation}
\vec{M} := \begin{pmatrix}
- \kappa & \frac{1}{c_* + \gamma} \\
\frac{1}{c_* + \gamma} & - \left( \kappa + \frac{s_i}{2 \gamma} \right)
\end{pmatrix}
\end{equation}
whose characteristic equation is
\begin{equation}
(\lambda + \kappa) \left( \lambda + \kappa + \frac{s_i}{2 \gamma} \right) - \frac{1}{(c_* + \gamma)^2} = 0 \ .
\end{equation}
Equivalently,
\begin{equation}
\lambda^2 + 2 \left[  \kappa + \frac{s_i}{4 \gamma}    \right] \lambda + \left[ \kappa^2 + \frac{\kappa s_i}{2 \gamma} - \frac{1}{(c_* + \gamma)^2} \right] = 0 \ .
\end{equation}
The solutions of this equation (i.e., the eigenvalues of $\vec{M}$) are
\begin{equation}
\begin{split}
\lambda^{\pm}_i &= - \left(  \kappa + \frac{s_i}{4 \gamma}    \right) \pm \sqrt{ \left(  \kappa + \frac{s_i}{4 \gamma}    \right)^2 + \frac{1}{(c_* + \gamma)^2} - \kappa^2 - \frac{\kappa s_i}{2 \gamma}  } \\
&= - \left(  \kappa + \frac{s_i}{4 \gamma}    \right) \pm \sqrt{ \frac{s_i^2}{16 \gamma^2}    + \frac{1}{(c_* + \gamma)^2} } \ .
\end{split}
\end{equation}
Note that $\lambda_i^-$ is always negative, and so corresponds to decay. The other root may be positive or negative depending on the size of the regularization strength $\kappa \geq 0$. Beyond a critical $\kappa$ value
\begin{equation}
\kappa_{\text{crit}} := \sqrt{ \frac{s_{\text{min}}^2}{16 \gamma^2}    + \frac{1}{(c_* + \gamma)^2} } - \frac{s_{\text{min}}}{4 \gamma}  \ ,
\end{equation}
where $s_{\text{min}}$ is the smallest of the $s_i$, all modes decay and one obtains only the trivial zero solution for all $u_i$ and $v_i$. For $\kappa < \kappa_{\text{crit}}$, at least some modes survive.

Assume in what follows that $\kappa < \kappa_{\text{crit}}$. Then at least some $\lambda_i^+$ will be positive, which means that $u_i$ and $v_i$ along those directions have components which \textit{grow} with time. In particular, since the eigenvectors of $\vec{M}$ are
\begin{equation}
\vec{q}^{\pm} := \begin{pmatrix}
1 \\
(c_* + \gamma) (\kappa + \lambda^{\pm}) \ ,
\end{pmatrix}
\end{equation}
the solution $\vec{z}(\tau) := ( u_i(\tau), v_i(\tau) )^T$ equals
\begin{equation}
\vec{z} = e^{\vec{M} \tau} \vec{z}_0 = e^{\vec{M} \tau} ( c_+ \vec{q}^+ + c_- \vec{q}^- )  =   c_+ \vec{q}^+ e^{\lambda^+_i \tau} + c_- \vec{q}^- e^{\lambda^i_- \tau}
\end{equation}
for some coefficients $c_+$ and $c_-$. 

\paragraph{Enforcing initial conditions.} We can determine $c_+$ and $c_-$ by choosing them so that $\vec{z}(0) = ( u_i(0), v_i(0) )^T$. This requires enforcing
\begin{equation}
\begin{split}
u_i(0) &= c_+ + c_- \\
v_i(0) &= c_+ (c_* + \gamma) (\kappa + \lambda^{+}) + c_- (c_* + \gamma)(\kappa + \lambda^{-}) \ .
\end{split}
\end{equation}
The first condition implies $c_- = u_i(0) - c_+$. The second implies
\begin{equation}
\begin{split}
\frac{v_i(0)}{c_* + \gamma} &= c_+  (\kappa + \lambda^{+}) + c_-(\kappa + \lambda^{-}) \\
&= c_+  (\kappa + \lambda^{+}) + [ u_i(0) - c_+] (\kappa + \lambda^{-}) \\
&= u_i(0) (\kappa + \lambda^-) + c_+ ( \lambda^+ - \lambda^-)
\end{split}
\end{equation}
so
\begin{equation}
\boxed{ \ c_+ = \frac{1}{\lambda^+_i - \lambda^-_i} \left[ \frac{v_i(0)}{c_* + \gamma} - u_i(0) (\kappa + \lambda^-_i) \right] \ } \ .
\end{equation}
Because the mode which \textit{grows} as opposed to \textit{shrinks} will dominate after enough time has passed, to good approximation we have
\begin{equation}
\begin{split}
u_i(\tau) &\approx e^{ \lambda^+_i \tau} c_+ = \frac{1}{\lambda^+_i - \lambda^-_i} \left[ \frac{v_i(0)}{c_* + \gamma} - u_i(0) (\kappa + \lambda^-_i) \right] e^{\lambda^+_i \tau} \\
v_i(\tau) &\approx e^{ \lambda^+_i \tau} c_+ (c_* + \gamma) (\kappa + \lambda^+)  =  \frac{(c_* + \gamma) (\kappa + \lambda^+) }{\lambda^+_i - \lambda^-_i} \left[ \frac{v_i(0)}{c_* + \gamma} - u_i(0) (\kappa + \lambda^-_i) \right] e^{\lambda^+_i \tau} 
\end{split}
\end{equation}
for moderate $\tau$. 

Note that $\lambda^+_i$ is a \textit{decreasing} function of $s_i$. This implies that $\vec{u}$ and $\vec{v}$ will tend to get amplified more quickly along less noisy directions. After enough time has passed, the coefficients along the least noisy direction will tend to dwarf the others, so to good approximation we have
\begin{equation}
\begin{split}
\vec{u}(\tau) &\approx \frac{u(\tau)}{2} \vec{P}  \left[  \frac{\vec{v}(0)}{c_* + \gamma} - \vec{u}(0) (\kappa + \lambda^-_{\text{min}})    \right]  \\
\vec{v}(\tau) &\approx \frac{v(\tau)}{2} \vec{P}  \left[   \frac{\vec{v}(0)}{c_* + \gamma} - \vec{u}(0) (\kappa + \lambda^-_{\text{min}})     \right]  
\end{split}
\end{equation}
after the alignment phase, where $\lambda^-_{\text{min}}$ is the negative eigenvalue associated with the least noisy direction (i.e., the smallest $s_i$, which we can call $s_{\text{min}}$). Here, $\vec{P}$ is a matrix that projects into the least noisy subspace (which may be of arbitrary dimensionality) and $u(\tau)$ and $v(\tau)$ are mode amplitudes whose precise form is given above. Equivalently, $\vec{u}(\tau)$ and $\vec{v}(\tau)$ align in the \textit{least noisy} state-space direction.

Define the quantity
\begin{equation}
\Delta := \sqrt{ \frac{s_{\text{min}}^2}{16 \gamma^2}    + \frac{1}{(c_* + \gamma)^2} } - \frac{1}{c_* + \gamma} \ ,
\end{equation}
which is an increasing function of $s_{\text{min}}$. $\Delta$ quantifies the gap between the noise-free solution and the noisy solution. In terms of it, we can write our answer as
\begin{equation}
\begin{split}
\vec{u}(\tau) &\approx u(\tau) \vec{P}  \left[  \frac{[ \vec{u}(0) + \vec{v}(0) ]}{2} +  \frac{(c_* + \gamma)}{2 } \left(\frac{s_{\text{min}}}{4 \gamma}  + \Delta(s_{\text{min}}) \right) \vec{u}(0)    \right]  \\
\vec{v}(\tau) &\approx v(\tau) \vec{P}   \left[  \frac{[ \vec{u}(0) + \vec{v}(0) ]}{2} + \frac{(c_* + \gamma)}{2 } \left(\frac{s_{\text{min}}}{4 \gamma}  + \Delta(s_{\text{min}}) \right) \vec{u}(0)    \right] \ .
\end{split}
\end{equation}

\subsection{W dynamics}

Given $\vec{u}$ and $\vec{v}$ dynamics, $\vec{W}$ dynamics are extremely simple. Because $\vec{u}$ and $\vec{v}$ are, to good approximation, collinear, we can write
\begin{equation}
\vec{u}(\tau) \approx u(\tau) \vec{u}_0 \hspace{1in} \vec{v}(\tau) \approx v(\tau) \vec{u}_0
\end{equation}
for some vector $\vec{u}_0$ (which is an eigenvector of $\vec{S}$ with eigenvalue $\sigma^2$) and scalar amplitudes $u(\tau)$ and $v(\tau)$. Eq. \ref{eq:align_app} then implies that
\begin{equation}
\frac{d}{d\tau} \vec{W} \approx  \left[ \frac{u(\tau) v(\tau)}{( c_* + \gamma)^2}  - \frac{\sigma^2 v(\tau)^2}{4 \gamma^2} \right] \vec{u}_0 \vec{u}_0^T \ ,
\end{equation}
or equivalently that
\begin{equation}
\vec{W}(\tau) \approx [ \gamma - c(\tau) ] \vec{u}_0 \vec{u}_0^T
\end{equation}
since these updates are all in the same direction. $c(\tau)$ is a scalar which determines the overall amplitude of $\vec{W}$. In other words, $\vec{W}$ becomes a rank-one matrix aligned in the same direction as $\vec{u}$ and $\vec{v}$.

\clearpage

\section{Derivation of low-dimensional effective dynamics}
\label{app:low_dim_deriv}

In this appendix, we derive the effective learning dynamics associated with the post-alignment phase. Recall from Appendix \ref{app:align} and Sec. \ref{sec:align} that the outcome of alignment has
\begin{equation}
\boxed{ \ \vec{u}(\tau) \approx u(\tau) \vec{u}_0 \hspace{0.5in} \vec{v}(\tau) \approx v(\tau) \vec{u}_0 \hspace{0.5in} \vec{W}(\tau) \approx [ \gamma - c(\tau) ] \vec{u}_0 \vec{u}_0^T \ }
\end{equation}
for some (assumed normalized) $\vec{u}_0$ which is an eigenvector of $\vec{S}$ with $\vec{S} \vec{u}_0 = \sigma^2 \vec{u}_0$. As we found earlier, this ought to be the \textit{least noisy} state-space direction. 

We will call this the \textit{alignment ansatz}. We want to show that dynamics are closed under this ansatz, in the sense that it suffices to track the time evolution of the scalars $u$, $v$, and $c$. We will also show that gradient descent dynamics on the full objective become equivalent to gradient descent dynamics on a simpler effective objective.

Note that, under the alignment ansatz,
\begin{equation}
f(t) = \vec{v}^T e^{\vec{A} t} \vec{u} = u(\tau) v(\tau) e^{- c(\tau) t}  \ .
\end{equation}
\subsection{Dynamics are closed under the alignment ansatz}

\paragraph{$\vec{u}$ dynamics.} General $\vec{u}$ dynamics have the form
\begin{equation}
\frac{d}{d\tau} \vec{u} =  \int_0^{\infty} [ f_*(t) - f(t) ] e^{- \gamma t} e^{\vec{W}^T t} \vec{v}  \ dt  - \kappa \vec{u} \ .
\end{equation}
Under our ansatz, this simplifies to
\begin{equation}
\begin{split}
\frac{d}{d\tau} u(\tau) \vec{u}_0 &=  \left[ \int_0^{\infty} [ f_*(t) - f(t) ] e^{- c(\tau) t} v(\tau) \ dt  - \kappa u(\tau) \right] \vec{u}_0 \\
&=  \left[ \int_0^{\infty} [ e^{- c_* t} - u(\tau) v(\tau) e^{- c(\tau) t} ] e^{- c(\tau) t} v(\tau) \ dt  - \kappa u(\tau) \right] \vec{u}_0 \\
&=  \left[ \frac{v(\tau)}{c_* + c(\tau)} - \frac{u(\tau) v(\tau)^2}{2 c(\tau)}   - \kappa u(\tau) \right] \vec{u}_0 \ .
\end{split}
\end{equation}
We indeed find that only the scalar $u(\tau)$ changes, as desired.

\paragraph{$\vec{v}$ dynamics.} General $\vec{v}$ dynamics have the form
\begin{equation}
\frac{d}{d\tau} \vec{v} = \int_0^{\infty} [ f_*(t) - f(t) ] e^{- \gamma t} e^{\vec{W} t} \vec{u}  -  e^{- 2 \gamma t} e^{ \vec{W} t} \vec{S} e^{ \vec{W}^T t} \vec{v}  \  dt - \kappa \vec{v} \ .
\end{equation}
Under the ansatz,
\begin{equation}
\begin{split}
\frac{d}{d\tau} v(\tau) \vec{u}_0 &= \left[ \int_0^{\infty} [ f_*(t) - f(t) ]  u(\tau) e^{- c(\tau) t}  - \sigma^2 v(\tau) e^{- 2 c(\tau) t}  \  dt - \kappa v(\tau) \right] \vec{u}_0 \\
&= \left[ \int_0^{\infty} [ e^{- c_* t} - u(\tau) v(\tau) e^{- c(\tau) t} ]  u(\tau) e^{- c(\tau) t}  - \sigma^2 v(\tau) e^{- 2 c(\tau) t}  \  dt - \kappa v(\tau) \right] \vec{u}_0 \\
&=  \left[ \frac{u(\tau)}{c_* + c(\tau)} - \frac{ [ u(\tau)^2 + \sigma^2 ] v(\tau)  }{2 c(\tau)}   - \kappa v(\tau) \right] \vec{u}_0  
\end{split}
\end{equation}
where $\sigma^2 := \vec{u}_0^T \vec{S} \vec{u}_0$ is the noise level in the alignment direction.

\paragraph{$\vec{W}$ dynamics.} General $\vec{W}$ dynamics have the form
\begin{equation}
\begin{split}
\frac{d}{d\tau} \vec{W} &= \int_0^{\infty} [ f_*(t) - f(t) ] e^{- \gamma t} \int_0^t  e^{\vec{W}^T t'} \vec{v} \vec{u}^T e^{\vec{W}^T (t - t')} \ dt' dt\\
&- \int_0^{\infty} e^{-2\gamma t} \int_0^t \  e^{\vec{W}^T t'} \vec{v} \vec{v}^T e^{\vec{W} t} \vec{S} e^{\vec{W}^T (t - t')} \ dt' dt  \ .
\end{split}
\end{equation}
Under the ansatz, 
\begin{equation}
\begin{split}
- \frac{d}{d\tau} c(\tau) \vec{u}_0 \vec{u}_0^T &= u(\tau) v(\tau) \left[ \int_0^{\infty} [ f_*(t) - f(t) ]  \ t e^{-  c(\tau) t} dt \right] \vec{u}_0 \vec{u}_0^T \\
&- \sigma^2 v(\tau)^2 \int_0^{\infty} t e^{-2 c(\tau) t} dt \vec{u}_0 \vec{u}_0^T   \ .
\end{split}
\end{equation}
Simplifying further,
\begin{equation}
\begin{split}
- \frac{d}{d\tau} c(\tau) \vec{u}_0 \vec{u}_0^T &= u(\tau) v(\tau) \left[ \frac{1}{[ c_* + c(\tau) ]^2} - \frac{u(\tau) v(\tau)}{4 c(\tau)^2} \right] \vec{u}_0 \vec{u}_0^T \\
&- \frac{\sigma^2 v(\tau)^2 }{4 c(\tau)^2} \vec{u}_0 \vec{u}_0^T   \ ,
\end{split}
\end{equation}
or
\begin{equation}
 \frac{d}{d\tau} c(\tau) \vec{u}_0 \vec{u}_0^T =  \left( \frac{ v(\tau)^2 [ u(\tau)^2 + \sigma^2 ]}{4 c(\tau)^2}  - \frac{u(\tau) v(\tau)}{[ c_* + c(\tau) ]^2} \right) \vec{u}_0 \vec{u}_0^T \ .
\end{equation}
In all cases, we indeed discover that the dynamics are closed under this ansatz. In summary, because the direction $\vec{u}_0$ never changes under gradient descent dynamics, it suffices to study the reduced system
\begin{equation} \label{eq:dyn_red_app}
\begin{split}
\frac{d}{d\tau} u(\tau) &=   \frac{v(\tau)}{c_* + c(\tau)} - \frac{u(\tau) v(\tau)^2}{2 c(\tau)} - \kappa u(\tau) \\
\frac{d}{d\tau} v(\tau) &=   \frac{u(\tau)}{c_* + c(\tau)} - \frac{v(\tau) [ \sigma^2  + u(\tau)^2 ]}{2 c(\tau)}  - \kappa v(\tau) \\
\frac{d}{d\tau} c(\tau) &=  \frac{v(\tau)^2 [ u(\tau)^2 + \sigma^2 ]}{4 c(\tau)^2}  -  \frac{u(\tau) v(\tau)}{[c_* + c(\tau) ]^2} \ .
\end{split}
\end{equation}

\subsection{Reduced dynamics are gradient descent on an effective objective}

We can make an additional theoretically interesting point about this ansatz: not only is it dynamically closed, but the above ODEs are equivalent to gradient descent on an \textit{effective} objective function. Fortunately, this objective is exactly what one obtains by naively substituting the ansatz into the original objective. 

Recall that the full objective equals
\begin{equation}
J(\vec{u}, \vec{v}, \vec{W}) =  \int_0^{\infty}    \frac{1}{2} \left[ f(t) - f_*(t) \right]^2 +  \frac{e^{- 2 \gamma t}}{2}  \vec{v}^T e^{\vec{W} t} \vec{S} e^{\vec{W}^T t} \vec{v} \   dt   + \frac{\kappa}{2} \Vert \vec{u} \Vert^2  + \frac{\kappa}{2} \Vert \vec{v} \Vert^2 \ .
\end{equation}
Substituting our ansatz, it becomes
\begin{equation}
\begin{split}
 & \   \frac{1}{2} \int_0^{\infty}  \left\{ \  \left[ u v e^{- c t}  - e^{-c_* t} \right]^2 + \sigma^2  v^2 e^{- 2 c t} \ \right\} \   dt + \frac{\kappa}{2} (u^2 + v^2)  \\
=& \   \frac{1}{2} \int_0^{\infty}  \left\{ \  u^2 v^2 e^{- 2 c t} - 2  u v e^{- ( c_* + c ) t} + e^{- 2 c_* t} + \sigma^2  v^2 e^{- 2 c t} \ \right\} \   dt + \frac{\kappa}{2} (u^2 + v^2)  \\
=& \  \frac{1}{2}   \left\{ \  u^2 v^2 \frac{1}{2 c} - 2  u v \frac{1}{ c_* + c} + \frac{1}{2 c_*} + \sigma^2 v^2 \frac{1}{2 c} \ \right\} + \frac{\kappa}{2} (u^2 + v^2)   \ .
\end{split}
\end{equation}
Simplifying, we obtain
\begin{align} \label{eq:J_eff_app}
J &=   \frac{u(\tau)^2 v(\tau)^2}{4 c(\tau)} -  \frac{u(\tau) v(\tau)}{ c_* + c(\tau) } + \frac{1}{4 c_*} +  \frac{\sigma^2 v(\tau)^2}{4 c(\tau)}  + \frac{\kappa}{2} (u^2 + v^2)   \\
&=    \frac{1}{4 c(\tau)} \left[  u(\tau) v(\tau) - \frac{2 c(\tau)}{c_* + c(\tau)} \right]^2 + \left( \frac{1}{4 c_*} - \frac{c(\tau)}{[ c_* + c(\tau) ]^2} \right) +  \frac{\sigma^2 v(\tau)^2}{4 c(\tau)} + \frac{\kappa}{2} [ u(\tau)^2 + v(\tau)^2 ]  \ . \notag
\end{align}
It is straightforward to verify that the above ODEs directly correspond to gradient descent on this objective, which establishes our claim.

\clearpage

\section{Characterization of amplitude-collapse phase transition}
\label{app:ps1}

In this appendix, we characterize the noise-induced phase transition associated with amplitude collapse (see Sec. \ref{sec:ps1}). The objective $J$ has the form
\begin{equation}
J =    \frac{1}{4 c} \left(  u v - \frac{2 c}{c_* + c} \right)^2 + \left( \frac{1}{4 c_*} - \frac{c}{( c_* + c )^2} \right) +  \frac{\sigma^2 v^2}{4 c} + \frac{\kappa}{2} ( u^2 + v^2 ) \ .
\end{equation}
This amounts to determining the conditions under which $J$ has a nontrivial optimum.

\subsection{Minimizing $J$ with respect to its arguments}

Let us begin by trying to optimize $J$ with respect to its arguments---although we will find it useful to work in terms of slightly different variables. Define the \textit{filter amplitude} $N := u v$ and the \textit{scale ratio} $r := u/v$. Since $v^2 = N/r$ and $u^2 = N r$, we can write $J$ as
\begin{equation}
J(r, N, c) =    \frac{1}{4 c} \left(  N - \frac{2 c}{c_* + c} \right)^2 + \left( \frac{1}{4 c_*} - \frac{c}{( c_* + c )^2} \right) +  \frac{\sigma^2 N}{4 c r} + \frac{\kappa}{2} N \left( \frac{1}{r} + r \right) \ .
\end{equation}

\paragraph{Minimizing with respect to $r$.} The derivative of $J$ with respect to $r$ equals
\begin{equation}
\frac{\partial J}{\partial r} = \frac{\kappa}{2} N - \frac{\kappa}{2} N \left(1 + \frac{\sigma^2}{2 \kappa c}  \right) \frac{1}{r^2} = 0
\end{equation}
which implies that
\begin{equation}
r^2 = 1 + \frac{\sigma^2}{2 \kappa c}
\end{equation}
at the minimum. Note that it depends on $c$ but not on $N$. In a slight abuse of notation, below we will use $r$ to denote the value of the scale ratio at the minimum. Without loss of generality, choose the positive solution so that
\begin{equation}
\boxed{ \ r = \sqrt{ 1 + \frac{\sigma^2}{2 \kappa c} } \ } \ .
\end{equation}
Since $r$ is defined as the ratio of $u$ and $v$, this immediately implies that $u$ and $v$ have the same sign at a nontrivial minimum. Relatedly, it must be true that $N \geq 0$, since $N = u^2 / r \geq 0$.

\paragraph{Minimizing with respect to $N$.} The derivative of $J$ with respect to $N$ equals
\begin{equation}
\begin{split}
\frac{\partial J}{\partial N} &= \frac{1}{2 c} \left( N - \frac{2 c}{c_* + c}  \right) + \frac{\kappa}{2} \left( 1 + \frac{\sigma^2}{2 \kappa c} \right) \frac{1}{r} + \frac{\kappa}{2} r  \\
&= \frac{1}{2 c} \left( N - \frac{2 c}{c_* + c}  \right) + \kappa r \ .
\end{split}
\end{equation}
Setting the left-hand side equal to zero, it appears true that
\begin{equation} 
N = \frac{2 c}{c_* + c} - 2 \kappa r c 
\end{equation}
at the minimum. But since our $r$ minimization implied that $N \geq 0$, we actually have
\begin{equation}
N = 2 c \left[ \frac{1}{c_* + c} - \kappa r \right]_+
\end{equation}
where $[ \cdots ]_+$ denotes rectification (i.e., a ReLU). That is, if the bracketed quantity is negative or zero, $J$ is minimized when $N = 0$. This is a sign of \textit{amplitude collapse}---if $N = 0$ is optimal, it is optimal to take $f(t) = 0$.

If we define the quantity
\begin{equation}
\boxed{ \ h(c) :=  \frac{1}{c_* + c} - \kappa r \ }  \ ,
\end{equation}
we can write the optimal $N$ as
\begin{equation}
\boxed{ \ N = 2 c [ h(c) ]_+ \ } \ .
\end{equation}

\paragraph{Rewriting the objective in terms of $N$.} In terms of $N$, the objective can be written at its minimum as
\begin{equation}
\begin{split}
J(c) &= c \left( \frac{1}{c_* + c} - [ h(c) ]_+  \right)^2 +  \frac{1}{4 c_*} - \frac{c}{( c_* + c )^2}   + 2 \kappa c [h(c)]_+ r(c) \\
&= \frac{1}{4 c_*} + c [h(c)]_+ \left( - \frac{2}{c_* + c} + [ h(c) ]_+  \right)  + 2 \kappa c [h(c)]_+ r(c) \\
&= \frac{1}{4 c_*} + c [h(c)]_+ \left( 2 \kappa r(c) - \frac{2}{c_* + c} + [ h(c) ]_+  \right)   \ .
\end{split}
\end{equation}
Either $h(c) \geq 0$, in which case the bracketed quantity equals $- h(c)$, or $h(c) < 0$, in which case the $[ h(c) ]_+$ prefactor makes the entire quantity zero. We conclude that
\begin{equation} \label{eq:J_red_app}
\boxed{ \ J(c) = \frac{1}{4 c_*} - c [h(c)]_+^2 }
\end{equation}
at the minimum. This immediately tells us that if $h < 0$ for all $c \geq 0$, there is a trivial solution with $N = 0$ and $J$ independent of $c$ (see Fig. \ref{fig:ps1}c).

\subsection{Determining the phase boundary}

While it appears prohibitively difficult to derive the exact $c$ that minimizes $J$ (since this requires solving a high-order polynomial equation), we can instead determine where there is a transition from a nontrivial solution to a trivial one.

\paragraph{Phase boundary conditions.} Eq. \ref{eq:J_red_app} implies that a nontrivial solution exists if and only if $h(c) > 0$ for some $c > 0$, since this yields $N > 0$ and a nontrivial value $c > 0$ at the minimum. On the other hand, if $h(c) \leq 0$ for all $c > 0$, then $N = 0$ and the trivial solution is optimal. The phase boundary therefore occurs when the maximum of $h(c)$ just touches zero. Hence, the phase boundary is characterized by the two conditions
\begin{equation}
\boxed{ \ h(c) = 0 \hspace{1in} h'(c) = 0  \ } \ .
\end{equation}
Explicitly, we want
\begin{equation}
\begin{split}
 \frac{1}{c_* + c} &= \kappa  \sqrt{1 + \frac{\sigma^2}{2 \kappa c}}  \\
  \frac{1}{(c_* + c)^2} &= \kappa \frac{\frac{\sigma^2}{2 \kappa c^2}}{2  \sqrt{1 + \frac{\sigma^2}{2 \kappa c}} } \ .
 \end{split}
\end{equation}
Relatedly, a nontrivial solution only exists when $c_* \kappa < 1$, since for $c_* \kappa \geq 1$, 
\begin{equation}
h(c) = \frac{1}{c_* + c} - \kappa r \leq \frac{1}{c_* + c} - \kappa < \frac{1}{c_*} - \kappa \leq 0 
\end{equation}
for all $c > 0$. In this regime, because $h(c)$ is always nonpositive, $\sigma^2_{\text{crit}} = 0$.

\paragraph{Some algebra.} It will be helpful to write $\sigma^2$ and $\kappa$ separately in terms of $c$. Divide the second equation by the first to obtain
\begin{equation}
\frac{1}{c_* + c} = \frac{\sigma^2}{4 \kappa c^2 r^2} \ .
\end{equation}
Rearranging, 
\begin{equation}
r^2 = 1 + \frac{\sigma^2}{2 \kappa c} = \frac{\sigma^2}{4 \kappa c^2} (c_* + c) \implies 1 + \frac{\sigma^2}{4 \kappa c} = \frac{\sigma^2}{4 \kappa c^2} c_* \ ,
\end{equation}
so
\begin{equation}
\frac{\sigma^2}{2 \kappa c} = \frac{2}{c_*/c - 1} \ .
\end{equation}
This implies that $r$ can be written in terms of $c$ as
\begin{equation}
r(c)^2 = 1 + \frac{2}{c_*/c - 1} = \frac{c_*/c + 1}{c_*/c - 1} = \frac{c_* + c}{c_* - c} \ .
\end{equation}
Using this result, the first condition can be rewritten as
\begin{equation}
\kappa^2 = \frac{1}{(c_* + c)^2 r^2} = \frac{c_* - c}{(c_* + c)^3}  \ .
\end{equation}
Let $x := c/c_*$ denote the non-dimensionalized learned rate. In terms of $x$, our two conditions read
\begin{equation}
\boxed{ \ \kappa^2 =  \frac{1}{c_*^2} \frac{1 - x}{(1 + x)^3}  \hspace{1in} \sigma^2 =  \frac{4 c_* \kappa x^2}{1 - x}  \ } \ .
\end{equation}
Since both $\sigma^2$ and $\kappa$ can be written in terms of the optimal value of $c$ (or equivalently, $x$), it suffices to determine this value. We can then write $\sigma^2$ in terms of $\kappa$, and hence determine the phase boundary.

\paragraph{Solving a cubic equation.} Note that the first condition expresses a cubic equation (in terms of $c_*$ and $\kappa$) whose solution is $x$:
\begin{equation}
c_*^2 \kappa^2 (1 + x)^3 = 1 - x \ .
\end{equation}
Note that it only has a valid solution if $0 \leq x < 1$. 

Generic cubic equations are annoying to solve, but they can be made tractable with a clever substitution. Consider the inspired choice
\begin{equation}
\boxed{ \ q := \sqrt{\frac{1 - x}{1 + x}}  \ } 
\end{equation}
which implies that
\begin{equation}
\boxed{ \ x = \frac{1 - q^2}{1 + q^2} \ } \ .
\end{equation}
Relatedly,
\begin{equation}
\begin{split}
1 + x &= \frac{2}{1 + q^2} \\
1 - x &= \frac{2 q^2}{1 + q^2} \ .
\end{split}
\end{equation} 
In terms of this new variable $q$, our cubic equation becomes
\begin{equation}
c_*^2 \kappa^2 = \frac{q^2}{4} (1 + q^2)^2 \ .
\end{equation}
Since $q \geq 0$ by definition, we can take the square root of both sides to find that
\begin{equation}
2 c_* \kappa = q (1 + q^2) \ ,
\end{equation}
or equivalently that
\begin{equation}
\boxed{ \ q^3 + q - 2 \kappa c_* = 0 \ } \ .
\end{equation}
This is a \textit{depressed cubic equation}. Its discriminant equals
\begin{equation}
\Delta = - 4 - 27 (- 2 \kappa c_*)^2 < 0 \ ,
\end{equation}
so it has one real root and two complex conjugate roots. Because a valid solution must be \textit{real}, we are interested in the real root. 

Using a standard formula for this root, we find that it equals
\begin{equation}
\boxed{ \ q(\kappa) = \frac{2}{\sqrt{3}} \sinh\left[ \frac{1}{3} \text{arsinh}(3 \sqrt{3} \kappa c_* ) \right] \ }
\end{equation}
where 
\begin{equation}
\text{arsinh}(x) := \log( x + \sqrt{x^2 + 1} ) \ .
\end{equation}

\paragraph{Determining the critical noise level.} Combining this result with our earlier formula for $\sigma^2$ in terms of $x$, 
\begin{equation}
\sigma^2 = \frac{4 c_* \kappa x^2}{1 - x} = 4 c_* \kappa \frac{(1 - q^2)^2}{2 q^2 (1 + q^2)}  \ .
\end{equation}
Since $2 c_* \kappa = q (1 + q^2)$, this becomes
\begin{equation}
\boxed{ \ \sigma^2_{\text{crit}} =   \frac{(1 - q(\kappa)^2)^2}{q(\kappa)} \ } \ .
\end{equation}

\paragraph{The final phase boundary result.}  Summarizing, at the boundary, provided $0 < \kappa c_* \leq 1$,
\begin{equation}
\begin{split}
q &= \frac{2}{\sqrt{3}} \sinh\left[ \frac{1}{3} \text{arsinh}(3 \sqrt{3} \kappa c_* ) \right] \\
\sigma^2 &=   \frac{(1 - q^2)^2}{q} \ .
\end{split}
\end{equation}

\clearpage

\section{Characterization of frequency-runaway phase transition}
\label{app:ps2}

In this appendix, we characterize the `frequency-runaway' phase transition discussed in Sec. \ref{sec:ps2} that occurs when $u$ and $v$ are frozen (with $u = v = 1$) but $c$ is trainable.

If $u$ and $v$ are frozen after alignment, the reduced objective becomes
\begin{equation}
J =   \frac{1 + \sigma^2}{4 c(\tau)} - \frac{1}{c_* + c(\tau)} + \frac{1}{4 c_*} 
\end{equation}
up to an additive constant. (Because it is constant, we have removed the regularization term). The negative gradient of $J$ equals
\begin{equation}
\begin{split}
-  \frac{\partial J}{\partial c} &= \frac{(1 + \sigma^2)}{4 c^2} - \frac{1}{(c_* + c)^2} \\
&= \frac{(1 + \sigma^2) (c_* + c)^2 - 4 c^2}{4 c^2 (c_* + c)^2} \\
&= \frac{[ \sqrt{1 + \sigma^2} (c_* + c) + 2 c] [ \sqrt{1 + \sigma^2} (c_* + c) - 2 c]}{4 c^2 (c_* + c)^2} \ ,
\end{split}
\end{equation}
so $c$ has dynamics
\begin{equation} 
\frac{d}{d\tau} c = \frac{[ \sqrt{1 + \sigma^2} (c_* + c) + 2 c] }{4 c^2 (c_* + c)^2} [ \sqrt{1 + \sigma^2} (c_* + c) - 2 c] \ .
\end{equation}
The fixed point of $c$ dynamics can only depend on the second factor in the numerator, since the first factor is always positive for $c \geq 0$ and $c_* > 0$. The second factor equals zero at
\begin{equation}
\boxed{ \ c_{\text{opt}} = \frac{\sqrt{1 + \sigma^2} \ c_*}{2 - \sqrt{1 + \sigma^2}}  \ }  \ ,
\end{equation}
so when $c_{\text{opt}}$ is well-defined, it is the fixed point of $c$ dynamics. 

Importantly, this quantity blows up when its denominator equals zero, which occurs at the critical noise level $\sigma^2_{\text{crit}} := 3$. Just below that noise level,
\begin{equation}
\boxed{ \ c_{\text{opt}} \approx \frac{8  c_*}{\sigma^2_{\text{crit}} - \sigma^2}  \ } \ ,
\end{equation}
so it diverges as $1/(\sigma^2_{\text{crit}} - \sigma^2)$. For $\sigma^2 \geq \sigma^2_{\text{crit}}$, the second factor equals
\begin{equation}
 \sqrt{1 + \sigma^2} (c_* + c) - 2 c = \sqrt{1 + \sigma^2} c_* + ( \sqrt{1 + \sigma^2} - 2) c \ ,
\end{equation}
which is always positive. This means that, beyond the critical noise level, dynamics drive $c$ to infinity. Hence, beyond the threshold, $J$ is a decreasing function of $c$ and has no finite minimum.

\end{document}